\pdfoutput=1
\documentclass[10pt,twocolumn,letterpaper]{article}

\usepackage{cvpr}
\usepackage{times}
\usepackage{epsfig}
\usepackage{graphicx}
\usepackage{amsmath}
\usepackage{amssymb}

\usepackage[table]{xcolor}
\usepackage{subcaption}
\usepackage{soul}
\usepackage{booktabs}
\usepackage[normalem]{ulem}
\usepackage{multirow}
\usepackage{pifont}% http://ctan.org/pkg/pifont

\newcolumntype{L}[1]{>{\raggedright\let\newline\\\arraybackslash\hspace{0pt}}m{#1}}
\newcolumntype{C}[1]{>{\centering\let\newline\\\arraybackslash\hspace{0pt}}m{#1}}
\newcolumntype{R}[1]{>{\raggedleft\let\newline\\\arraybackslash\hspace{0pt}}m{#1}}

\newif\ifdraft
\draftfalse
\drafttrue

\usepackage[table]{xcolor}
\definecolor{orange}{rgb}{1,0.5,0}
\definecolor{violet}{RGB}{70,0,170}

\ifdraft
 \newcommand{\PF}[1]{{\color{red}{\bf PF: #1}}}
 
 \newcommand{\MS}[1]{{\color{green}{\bf MS: #1}}}
 
 \newcommand{\ZD}[1]{{\color{violet}{\bf ZD: #1}}}
 
 \newcommand{\YH}[1]{{\color{blue}{\bf YH: #1}}}
 
 \newcommand{\SPE}[1]{{\color{orange}{\bf SS: #1}}}
 
 \newcommand{\WJ}[1]{{\color{orange}{\bf WJ: #1}}}
 
\else
 \newcommand{\PF}[1]{}
 
 \newcommand{\KY}[1]{}
 
 \newcommand{\MS}[1]{}
 
 \newcommand{\ZD}[1]{}
 
 \newcommand{\YH}[1]{}
 
 \newcommand{\SPE}[1]{}
 
 \newcommand{\WJ}[1]{}
 
\fi

\newcommand{\comment}[1]{}

\newcommand{\bK}{\mathbf{K}}
\newcommand{\bR}{\mathbf{R}}

\usepackage[pagebackref=false,breaklinks=true,letterpaper=true,colorlinks,bookmarks=false]{hyperref}

\cvprfinalcopy % *** Uncomment this line for the final submission

\def\cvprPaperID{3328} % *** Enter the CVPR Paper ID here

\ifcvprfinal\fi

\begin{document}

%%%%%%%%% TITLE
\title{Wide-Depth-Range 6D Object Pose Estimation in Space}

\author{%
	{Yinlin Hu $^1$, \quad S\'ebastien Speierer $^2$, \quad Wenzel Jakob $^2$, \quad Pascal Fua $^1$, \quad Mathieu Salzmann $^{1,3}$} \\
	{\small $^1$ EPFL Computer Vision Lab, \quad $^2$ EPFL Realistic Graphics Lab, \quad $^3$ ClearSpace SA} \\
	{\tt \small \{firstname.lastname\}@epfl.ch}\\
}

\maketitle

% !TEX root = ../top.tex
% !TEX spellcheck = en-US

%%%%%%%%% ABSTRACT
\begin{abstract}
% Performing reliable 6D pose estimation in space poses challenges that most earth-bound algorithms do not address. Chief among them is the fact that objects can be seen at very different scales, a problem that is underrepresented in traditional benchmark datasets and, hence, underappreciated in our community.
%
% Existing approaches to solving this problem rely on a two-stage approach that first estimates scale and then pose on a resized image patch. We propose instead a single-stage hierarchical end-to-end trainable network that yields better robustness to scale variations.
% We will demonstrate that it outperforms existing approaches not only on images synthesized to resemble images taken in space but also on standard benchmarks.

    6D pose estimation in space poses unique challenges that are
    not commonly encountered in the terrestrial setting. One of the most
    striking differences is the lack of atmospheric scattering, 
    allowing objects to be visible from a great distance while complicating illumination conditions.
    %which implies
    %that objects can be observed at vastly different scales as the camera (e.g.,
    %on a satellite) approaches an object. \MS{Why is this due to the lack of atmospheric scattering?} 
    %\YH{In my understanding, this is because the air near the earth's surface can also occlude something?
    %While, to be honest, this is too technical.}
    Currently available benchmark
    datasets do not place a sufficient emphasis on this aspect and mostly
    depict the target in close proximity.

    Prior work tackling pose estimation under large scale variations relies on a two-stage approach to first estimate
    scale, followed by pose estimation on a resized image patch. We instead
    propose a single-stage hierarchical end-to-end trainable network that is
    more robust to scale variations. We demonstrate that it outperforms
    existing approaches not only on images synthesized to resemble images taken
    in space but also on standard benchmarks.
\end{abstract}

%Earth orbits become more and more congested, which is a severe threat to space safety. Estimating 6D pose of space-borne objects is the crucial component of the future solutions, such as debris removal and spacecraft interactivity. Unlike general 6D object pose estimation, the object in space often exhibits a wide depth range, making the general method suffering from the scaling problem. A straightforward strategy is to detect the target first and resize the detected object to a uniform resolution before pose estimation, which, however, is heavy and suboptimal. In this paper, we propose to use a single hierarchical network to handle this problem.  Moreover, we introduce a scale-aware sampling strategy during training to make pyramid levels interactive with others. Based on it, we show that pose fused from multiple levels can achieve much more accurate results. To fully demonstrate the effectiveness of our method in space, we introduce a photorealistic satellite dataset based on a Cubesat-type satellite SwissCube, which is the first satellite dataset with accurate 3D models, complex movement modelings, and also physical simulations of the Sun, Earth, and galaxies. Furthermore, we show that our multi-scale fusion framework is also effective in the general 6D object pose estimation, demonstrated by achieving state-of-the-art performance on the general 6D pose dataset.

% !TEX root = ../top.tex
% !TEX spellcheck = en-US

\section{Introduction}
\label{sec:introduction}

\begin{figure}[t]
\centering
\includegraphics[height=2.2cm]{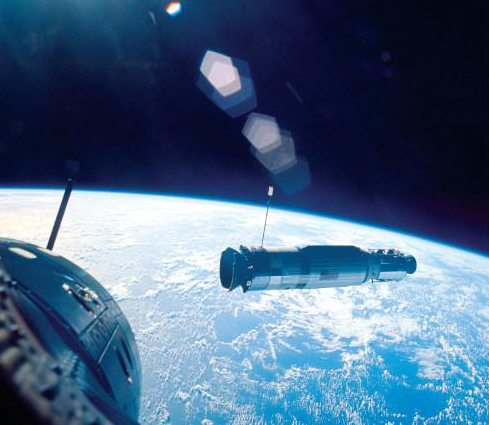}
\includegraphics[height=2.2cm]{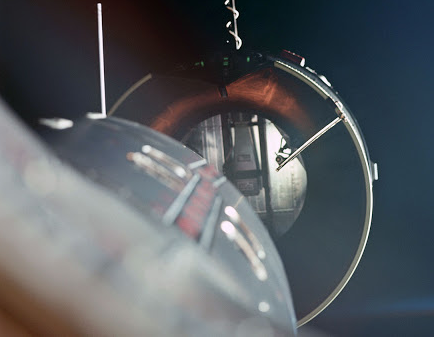}
\includegraphics[height=2.2cm]{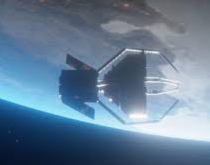}
\begin{small}
\begin{tabular}{C{0.3\linewidth}C{0.2\linewidth}C{0.4\linewidth}}
(a)&(b)&(c)
% (c)&(d)
\end{tabular}
\end{small}
\vspace{-6mm}
\caption{\small {\bf Docking and space cleaning.} {\bf (a, b)} Two different views of the Agena target vehicle during the first space docking. The appearance of Agena is strongly affected by the large scale and viewpoint changes, suggesting that different image features should be used for 6D pose estimation. In 1966, this docking procedure was controlled manually. {\bf (c)} In 2025, the ClearSpace One chaser satellite will be launched to retrieve and de-orbit a non-operational satellite, so as to showcase the feasibility of removing space debris. In this case, the capture will be fully automated. The synthetic image shown here highlights the challenges the algorithm will have to handle, such as reflections, over-exposure of some parts of the images, and lack of details in others. 
% \MS{I would tend to use the current image (d) in (c), and another image better showing these challenges, and possibly at a different scale, here.} \PF{Aren't images (a) and (b) challenging enough? You are looking at the same target in both.}
% \YH{Fixed, the main meanings are still the same but save more space.}
}
\label{fig:docking}
\end{figure}

Reliable 6D pose estimation is key to automating many spatial maneuvers, such
as docking or capturing inert objects as shown in Fig.~\ref{fig:docking}. An
important consequence of such maneuvers is that they dramatically change the
scale and aspect of the observed target. Although 6D pose estimation is an
active area of research in computer vision and robotics, this important aspect
has not received significant attention thus far---for example, most benchmark
datasets~\cite{Hinterstoisser12b,Krull15,Xiang18b,Hodan18} feature objects whose depth varies within a limited range.
The lack of atmospheric scattering enabling observation from great
distances also leads to other challenges: harsh contrast, under- and
over-exposed areas, and significant specular reflections from reflective
materials used in space engineering (aluminium and carbon fiber panels, etc.).
% \WJ{(Many edits here, and in the abstract -- remove this comment, if they are okay.)}
% \YH{Seems great}

 % !TEX root = ../top.tex
% !TEX spellcheck = en-US

To address such challenges, the European Space Agency (ESA) and Stanford
University recently organized a satellite pose estimation challenge based on
the \emph{Spacecraft Pose Estimation Dataset} (SPEED)~\cite{Kisantal20}. The best-performing
methods in this competition use a two-step approach to handle large depth
variation: a detector finds an axis-aligned box bounding the target, which is
resampled to a uniform size and finally processed by a 6D pose estimator.

\begin{figure}[t]
\centering
\includegraphics[width=0.49\linewidth]{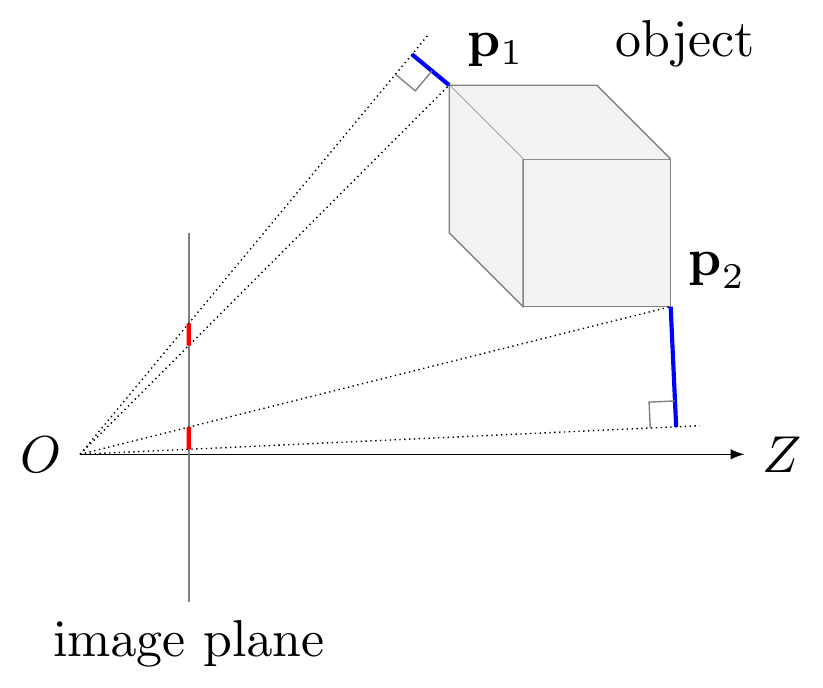}
\includegraphics[width=0.49\linewidth]{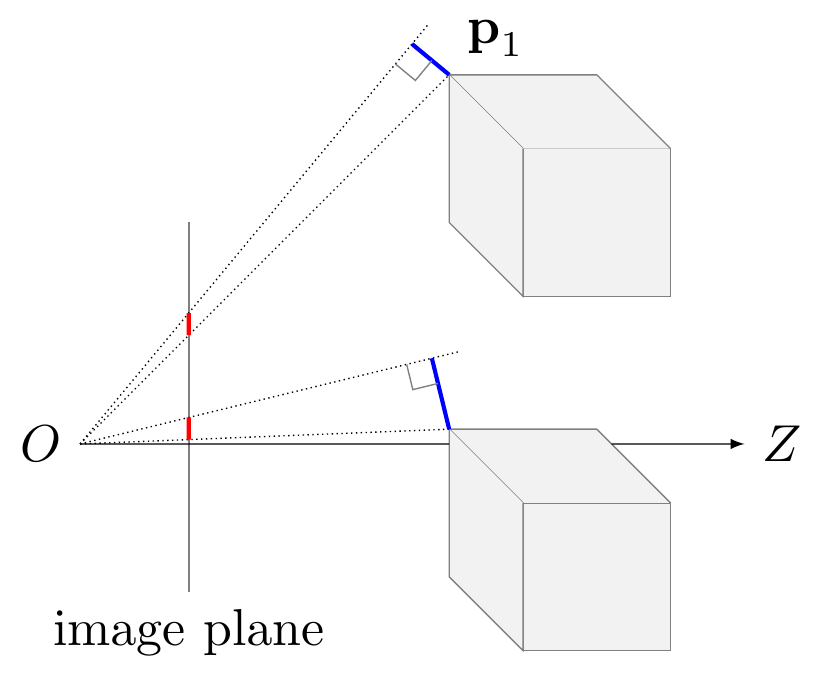}
\begin{tabular}{C{0.45\linewidth}C{0.45\linewidth}}
{\scriptsize (a) Sensitivity to different keypoints} & {\scriptsize (b) Sensitivity to target positions}
\end{tabular}
\vspace{-5mm}
\caption{\small {\bf Problem with minimizing the 2D reprojection error.} {\bf (a)} The red lines denote the 2D reprojection errors for points ${\bf p}_1$ and ${\bf p}_2$. Because one is closer to the camera than the other, these 2D errors are of about the same magnitude even though the corresponding 3D errors, shown in blue, are very different. {\bf(b)} For the same object at different locations, the same 2D error can generate different 3D errors. This makes pose accuracy dependent on the relative position of the target to the camera. 
% The problem depicted by (a) has little impact as long as the size of the target is small with respect to its distance to the camera but worsens when the object comes closer. The exact opposite occurs in the case depicted by (b). \YH{these statements may be not very accurate, so I removed them to avoid potential criticizes.}
% \WJ{Lower figure shows $\mathbf{p}_1$ twice. This figure consumes a relatively large amount of space for the content, you could probably show (a) and (b) side by side if space is needed.}\YH{Fixed}
}
\label{fig:cube_problem}
\end{figure}

This approach is suboptimal in several ways. First, detection and pose
estimation are treated as separate processes, which precludes joint training.
Second, it provides supervisory signals only to the final layer of the
encoder-decoder architecture being used instead of to all levels of the
decoding pyramid, which would increase robustness. Third, many similar feature
extraction computations are performed by both processes, which results in an
unnecessary duplication of effort. Finally, these methods rely on the dominant approach
to deep learning based 6D object pose estimation~\cite{Rad17,Hu19a,Chen19DLR} consisting of
training a network to minimize the 2D reprojection error of predefined 3D
keypoints, which cannot cope with large depth range variations: As shown
in Fig.~\ref{fig:cube_problem}, reprojection error is strongly affected by the
distance of individual keypoints to the camera, and not explicitly taking this
into account degrades performance. 
% \WJ{Use of ``keypoint'' in this paragraph seems unnecessarily technical, why not ``position''? Also applies to Fig 2 caption.}\YH{It may be confused with other points on the object surface. We only use the 8 corners of the 3D object bounding box of the object. So we say they are ``key'' points. And most literature uses this terminology.}

\begin{figure}[t]
\centering
\includegraphics[width=0.99\linewidth]{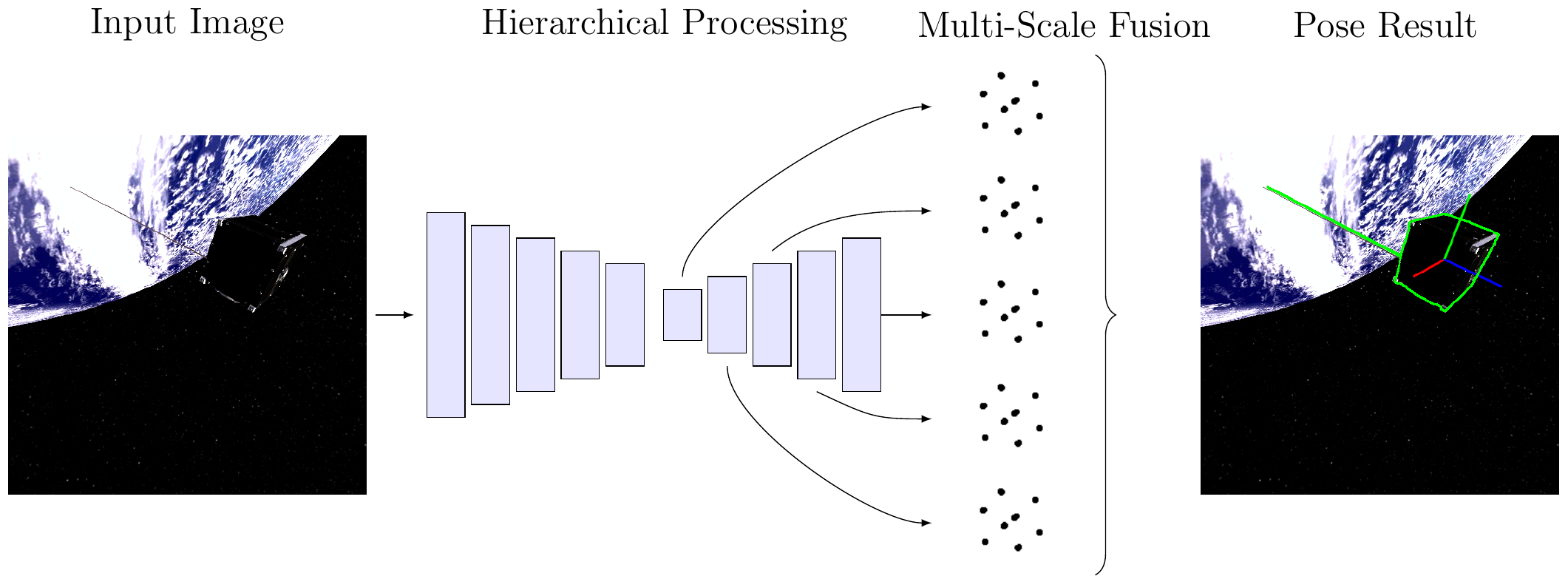}
\vspace{-3mm}
    \caption{{\bf Our single-stage approach.} We use an encoder-decoder architecture to progressively downsample the image and then to re-expand it. At each level of the decoder, we establish 3D-to-2D correspondences. Finally, we use a RANSAC-based PnP strategy~\cite{Lepetit09} 
    % \WJ{citation?} \YH{Fixed} 
     to infer a single reliable pose from these sets of correspondences. 
    % \WJ{Impossible to read text in printout, please try to have text in the figure with a similar font size as the surrounding article.}\YH{Fixed}
    }
\label{fig:arch}
\end{figure}

To address these shortcomings, we introduce a single hierarchical end-to-end trainable network depicted by Fig.~\ref{fig:arch} that yields robust and scale-insensitive 6D poses. 
To use information across scales, it progressively downscales the learned features,  derives 3D-to-2D correspondences for each level of the resulting pyramid, and finally uses a RANSAC-based PnP strategy to infer a single reliable pose from these sets of correspondences. This is a departure from most networks that estimate pose only from the final layer. 
To address the issue in Fig.~\ref{fig:cube_problem}, we minimize a training loss based on 3D positions instead of 2D projections, making the method invariant to the target distance.
We use a Feature Pyramid Network (FPN)~\cite{Lin17e} as our backbone but, unlike in most approaches relying on such networks, we assign each training instance to multiple pyramid levels to promote the joint use of multi-scale information. 

In short, our contribution is a new 6D pose estimation architecture that reliably handles large scale changes under challenging conditions. We will show that it outperforms all state-of-the-art methods on the established SPEED dataset while also being much faster. Furthermore, we introduce a larger-scale satellite pose estimation dataset featuring more realistic and more complex images than SPEED, and we show that our method delivers the same benefits in this more challenging scenario. Finally, we demonstrate that our method outperforms the state of the art even on images with smaller depth variations, such as those of the challenging Occluded LINEMOD dataset. Our code and new dataset will be publicly released.

% !TEX root = ../top.tex
% !TEX spellcheck = en-US

% !TEX root = ../top.tex
% !TEX spellcheck = en-US

%\PF{Based on our zoom meeting, I would recommend showing 
%\begin{itemize}
% \item quantitative results on SPEED. 
% \item qualitative results on VESPA to match Fig. 1. 
% \item qualitative results on real images of the space cube with a discussion about doing better in future work using DA. 
%\end{itemize}
%}
%
%\YH{Thank you. I will do it asap.}

%-------
% OLD
%-------

%Estimating 6D object pose has became an essential component of many real-world vision applications, including robotics,  machine perception, and augmented reality etc~\cite{xx}. On the other hand, estimating 6D pose of space objects, such as satellites and orbit debris, is drawing significant attention as the increasing congestion in Earth orbits~\cite{xx}.

%In a typical wide-depth-range scenario, estimating 6D object pose will be much more challenge due to the scaling problems as the estimation of farther samples will be dominated by closer ones~\cite{xx}. The most straightforward way to handle this probelm is to introduce a object detection network as a preprocessing component, and scale all the detected bounding boxes to the same size before feeding them into another pose regression network. However, this type of strategy equipping with two separated networks is heavy and suboptimal in practice~\cite{xx} (see Fig.~\ref{fig:pyramid_vs_twonetworks}).

% !TEX root = ../top.tex
% !TEX spellcheck = en-US

\section{Related Work}
\label{sec:related}

%Mounting a depth sensor or LiDAR onto a spacecraft is still not very common due to economic, stability, or power consumption reasons. In this work, we will focus on RGB-based 6D object pose estimation.
The most commonly-used sensors for 6D pose estimation in space remain cameras, may they be RGB, monochromatic, or, although more rarely, infrared. 
% \WJ{That seems potentially dubious without presenting any evidence. I could imagine that a faster monochromatic sensor may preferable sometimes.
% How about saying that you focus on RGB data, but the method works in principle for any kind of image-based sensor including monochromatic
% visible and infrared cameras?} \MS{Correct. In fact the original statement contrasted cameras with LiDAR, and my edit lost this point. I rephrased.}
We therefore focus on image-based 6D pose estimation in both our work and the discussion below.

The standard framework to perform 6D pose estimation consists of first establishing 3D-to-2D correspondences, and then compute the pose using a PnP solver~\cite{Lu00,Tulsiani15,Pavlakos17a}. While many hancrafted methods have been designed to extract the required correspondences~\cite{Lowe04,Tola10,Trzcinski12c}, they tend to produce low-quality output under challenging conditions (objects lacking spatial variation, strong highlights, etc.). As such, most modern 6D object pose estimation methods establish such correspondences using a neural network. This network is usually trained to predict the image location of the 3D object bounding box corners, either in a single global fashion~\cite{Kehl17,Rad17,Tekin18a,Xiang18b}, or by aggregating multiple local predictions to improve robustness to occlusions~\cite{Oberweger18,Jafari18,Hu19a,Peng19a,Zakharov19a,Li19a}. 
%Segmentation-driven 6D object pose estimation~\cite{Hu19a} first claim that local predictions from the feature cell located within the segmentation mask can be fused to a single robust 6D pose results. PVNet~\cite{Peng19a} adopts a similar manner while using a different representation of local predictions. Most of these methods only consider the case in a fixed image scale and can not handle the cases with apparent scale drastically changed. Our method pushes the limit of this type of method to wide-depth-range scenarios in a single compact hierarchical framework. 
Whether global or local, these methods were designed to be effective on standard computer vision benchmarks, which feature minimal scale changes. As we will show in our experiments, they therefore perform poorly when the depth range at which the object is depicted varies dramatically across different images.

The few works that have attempted to handle the scale issue rely on an object detection network as a preprocessing component~\cite{Li18a,Li19a,Chen19DLR}. While the zoom sampling strategy introduced in~\cite{Li19a} aims to account for the object detection noise when training the pose network, it still does not reflect the true distribution of the patches output by the detection network, and the resulting framework does not unify the detection and pose estimation stages.  While this could in principle be achieved via a Spatial Transformer Network~\cite{Jaderberg15}, such a change would significantly complicate the architecture, introducing redundant operations across the detection and pose estimation modules and eventually precluding real-time inference. Our main contribution entails using the inherent hierarchical structure of a single network with shared weights across the levels to handle the scale problem. We demonstrate this to be both robust and efficient.

Hierarchical processing, such as image pyramids~\cite{Bartoli08,Hu16,Jing18}, is a classical idea for multi-scale image understanding~\cite{Hu16a,Hu18d}. Recently, this idea has been translated to the deep learning realm via Feature Pyramid Networks (FPNs)~\cite{Lin17e}, which are now a standard component of many object detection frameworks~\cite{Lin17f,Tian19b,Zhang20d}. Here, we leverage this idea for 6D object pose estimation. However, unlike most object detection methods that explicitly associate each pyramid level to a single, predefined scale, we introduce a dynamic sampling strategy where each training instance leverages all pyramid levels, albeit with different weights. This allows us to fuse the predictions from the different levels at inference, leading to more robust 6D pose estimates.

 % !TEX root = ../top.tex
% !TEX spellcheck = en-US

\begin{figure}[t]
\begin{minipage}{\linewidth}
\centering
\includegraphics[height=2.2cm,trim=450 350 380 50,clip]{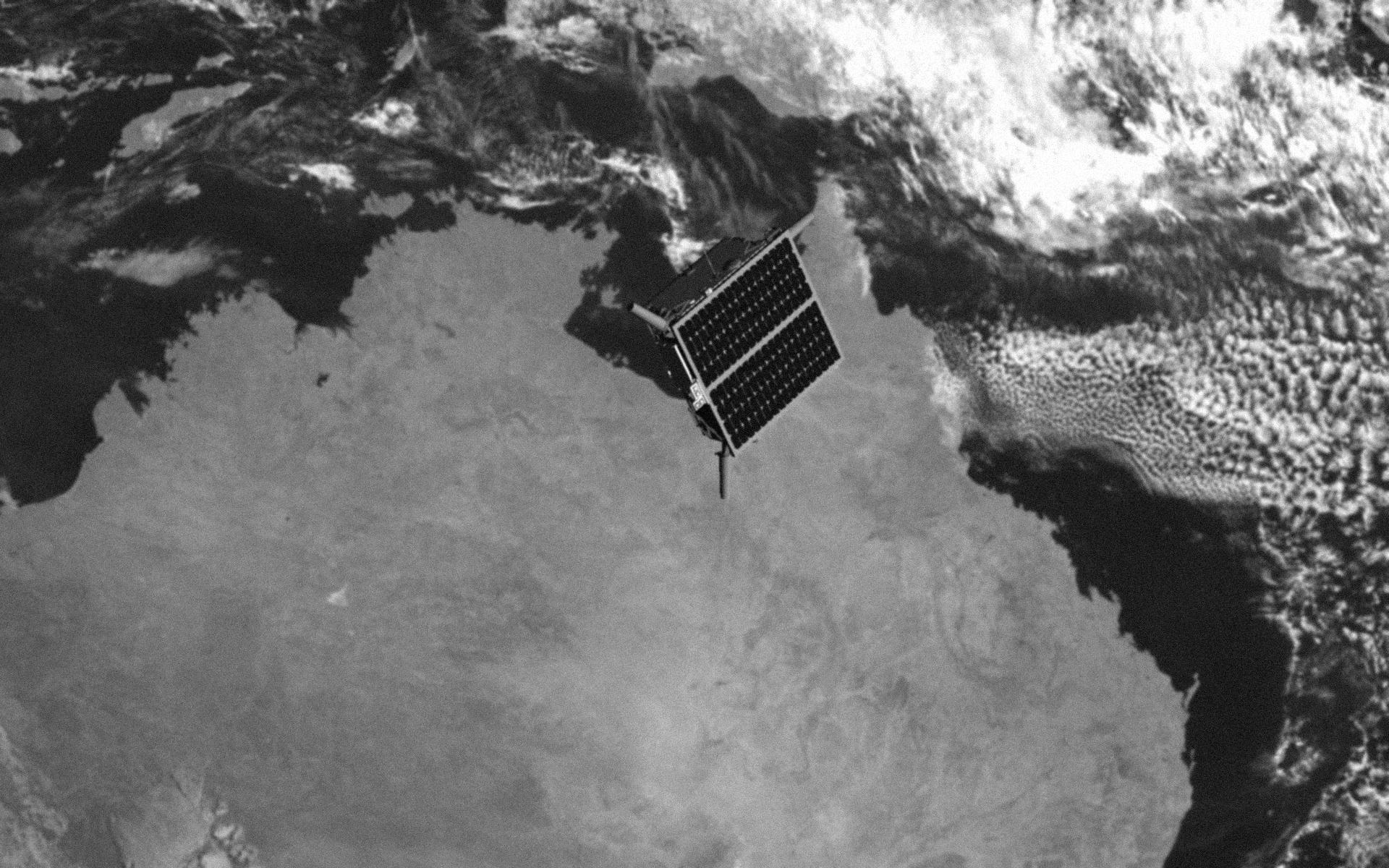}
\hspace{1em}
\includegraphics[height=2.2cm,trim=300 450 350 400,clip]{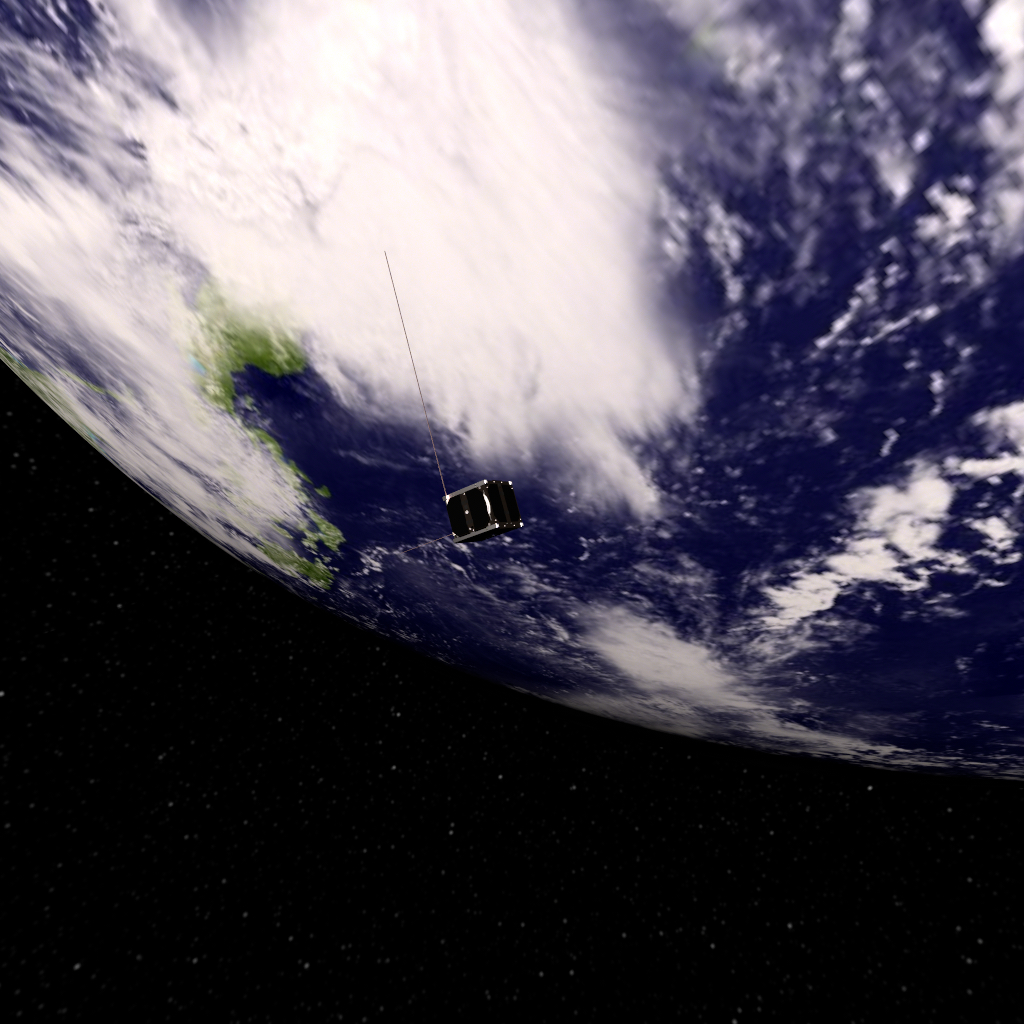}
\begin{scriptsize}
\begin{tabular}{C{0.35\linewidth}C{0.6\linewidth}}
(a) The SPEED dataset & (b) The proposed SwissCube dataset
% (c)&(d)
\end{tabular}
\end{scriptsize}

\vspace{-6mm}
\caption{\small {\bf Comparison of datasets.} {\bf (a)} The SPEED dataset~\cite{Kisantal20} was generated with a non-physics-based renderer and only poorly reflects the complexity of illumination in space. {\bf (b)} We introduce a SwissCube dataset that was created via physics-based rendering. 
%The dataset name is temporarily anonymized as CubeSat for not revealing authorship.
}
\label{fig:swisscube_vs_speed_demo}
\end{minipage}
\end{figure}

We focus our experiments on 6D pose estimation of space-borne objects, because robustness to scale is highly important in that context, particularly when approaching non-cooperative targets (e.g. space trash)
% \WJ{Removed the word "non-cooperative" here and in the intro. It sounds too much like the object actively fights being approached :-)} \MS{But this is the term commonly used in the space engineering literature.}\YH{Fixed}
that require motion synchronization. The space engineering community has its own literature on the topic of 6D pose estimation. While it has evolved in a manner that resembles progress in computer vision, it has mostly focused on handcrafted methods~\cite{Zhang05b,Amico14,Petit11,Sharma18b}, with only a few works proposing deep learning based approaches~\cite{Chen19DLR}. The main reason for this is the lack of large amounts of annotated data for space-borne objects. 
Recently, this was addressed by the SPEED dataset~\cite{Kisantal20} released by ESA and Stanford University as part of a satellite pose estimation challenge. This dataset, however, has several limitations. First, it does not provide the 3D model of the satellite, and while it can be reconstructed from the images, the final pose estimate will depend not only on the pose estimation algorithm but also on the quality of this reconstruction. 
% \WJ{It's not immediately apparent why that is a problem, other than maybe wanting to generate more data? It would be useful to have the model for a differentiable rendering-based optimization to improve an initial guess by a neural network, but that is probably beyond the scope of this discussion.} \YH{This is not related to the scale of the data, but to fair comparisons. Without an accurate 3D model, the evaluation of the 6D object pose will depend on many other factors, especially how you reconstruct the 3D model yourself. So we need a dataset with released accurate 3D model to get rid of these prerequisites.} 
Second, the SPEED images were synthesized by a non-physics-based rendering technique, only poorly reflecting the complexity of illumination in space, as illustrated in Fig.~\ref{fig:swisscube_vs_speed_demo}. 
% \WJ{is there something more concrete that could be said about this lack of realism? I would say that light transport in space is actually *much* easier than on earth, because there is no atmospheric scattering, and most directions are simply black. In some sense, even realistically generated space-based images look "fake" :-)} \YH{Have added a new figure}
% Finally, the SPEED dataset remains of relatively small size, with only XXX training images. \MS{Complete.} 
Finally, the depth distribution of the SPEED dataset is not uniform, with only few images depicting the satellite at a large distance from the camera. However, accurate pose for farther objects can be critical for space rendezvous; they give the docker or chaser enough time to adjust its own motion and prepare for the actual operations.
% and do other preparations at the very first time. 
% \WJ{Beginning of this sentence doesn't parse for me. First time a "grabber" is mentioned in this paper, should this occur earlier?}\YH{rephrased.}
We propose a novel satellite pose estimation dataset that addresses this bias,
and constitutes the second contribution of this article. The images in
this dataset were created using a physically-based spectral light transport simulation
involving an accurate reference 3D model of a cube satellite that accounts for
the effects of the Sun, Earth, stars, etc.

% !TEX root = ../top.tex
% !TEX spellcheck = en-US

\section{Approach}
\label{sec:approach}

Our goal is to estimate the 3D rotation and 3D translation of a known rigid object depicted in an RGB image. To this end, 
%following the recent trend~\cite{xx,Hu19}\WJ{Hm, "following the trend" doesn't sound like the best motivation, would omit. Better to argue NN yields unparalleled accuracy compared to classical vision approaches?}, 
we design a deep network that regresses the 2D projections of predefined 3D points. However, rather than regressing the 2D projections at a single, fixed scale, which lacks robustness to large depth variations, we use a Feature Pyramid Network (FPN)~\cite{Lin17e}, perform the regression at multiple scales, and fuse the resulting multiple estimates in a robust pose prediction.

In the following sections, we first present the FPN architecture our network builds on and then introduce a sampling-based training strategy to leverage every pyramid level for each training instance. Finally, we discuss our fusion approach to obtaining a single pose estimate during inference.

\subsection{Pyramid Network Architecture}
% \WJ{High-level impression: I had to look at the original paper at this point, because the
% discussion below didn't work for me as a self-contained
% explanation of FPN. While reading that, I was wondering:
% are you mainly building on the high-level idea, or is it a faithful reimplementation
% of the various lateral and top-down connections (which aren't mentioned anywhere). 
% You may want to add a sentence like "Our FPN architecture is identical to the original paper, except ...."
% The text discussing backbone network, strides etc., is very technical, and the main question should be: why did you make these modifications?
% "Objectness indicator", "thin branch layer", etc -- all of these changes raise similar questions that aren't explained.
% It may be helpful to structure the discussion by adding a figure of classic FPN and the modified architecture (perhaps with new blocks in a different color)?
% The dimensionality of feature vectors seems like a low level detail (perhaps add a detailed network description in the supplemental?).
% }
% \YH{rephrased. We do not invent a network architecture and, basically, use the FPN directly. Our point is how to train it to make it work on 6D object pose, as discussed in the following section. ensemble-aware sampling.}

Most 6D pose estimation deep networks rely on an encoder-decoder architecture. Therefore, to handle large scale variations for 6D object pose estimation, instead of relying on an additional object detection network, we use the inherent hierarchical architecture of the encoder network, which extracts features at different scales. Specifically, we use Darknet-53~\cite{Redmon18} as backbone in our framework and employ the same network architecture as in the FPN~\cite{Lin17e} designed for object detection, which consists of $k=5$ levels of feature maps, $\{{\cal F}_1, {\cal F}_2, {\cal F}_3, {\cal F}_4, {\cal F}_5\}$, each with an increasingly large receptive field. 
% The spatial width and height of the feature maps is then halved in each subsequent pyramid level. 
    % Following~\cite{RetinaNet, FCOS, ATSS}, the highest-resolution feature map ${\cal F}_1$ is obtained by using a stride of $8$ on the features from the third block of the backbone \yh{Darknet-53} network~\cite{1}. 

% The feature maps ${\cal F}_1, {\cal F}_2, {\cal F}_3, {\cal F}_4$ and ${\cal F}_5$ thus have strides $8, 16, 32, 64$ and $128$, respectively. \MS{I am not sure to understand this statement and why we need it.}

Instead of computing a single pose estimate from the feature map ${\cal F}_5$ only, we regress the 2D locations of the object 3D keypoints from every level of this pyramid. 
% \MS{How is this done? The different feature maps have different resolutions, so they cannot be processed by the same decoder. This needs some explanations. Check what I wrote thereafter.} \YH{We add a thin branch layer to each pyramid level of the same decoder to perform the final prediction for each level.} 
To this end, we rely on the segmentation-driven approach of~\cite{Hu19a}, and make the feature vector at every spatial location in each feature map to output the 2D projections of the 3D keypoints, represented as an offset from the center of the corresponding cell, and an objectness score for each object class. The feature vector at each cell therefore is a $C\times (2\times 8+1)$ dimensional vector consisting of $8$ 2D offsets and an objectness indicator for $C$ object classes. To encode a segmentation mask, all feature cells need to be involved in the objectness prediction, including those that contain no target objects. By contrast, as discussed below, only selected cells are involved in training the pose regressor.

%Most practical networks under the encoder-decoder structure, including most 6D object pose estimation networks~\cite{xx}, have a hierarchical architecture which extracts the features of different scalings level by level~\cite{xx}. 
%We propose to make use of those multiple feature levels that are already there rather than regressing the pose only on the last feature level which suffers from the scaling problem.

% !TEX root = ../top.tex
% !TEX spellcheck = en-US

\begin{figure}[t]
\centering
\includegraphics[width=0.42\linewidth]{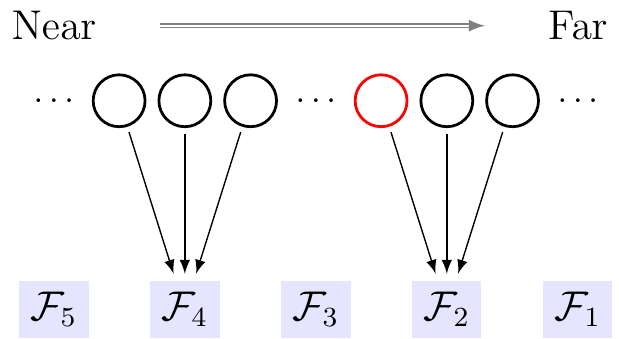}
\hspace{2em}
\includegraphics[width=0.42\linewidth]{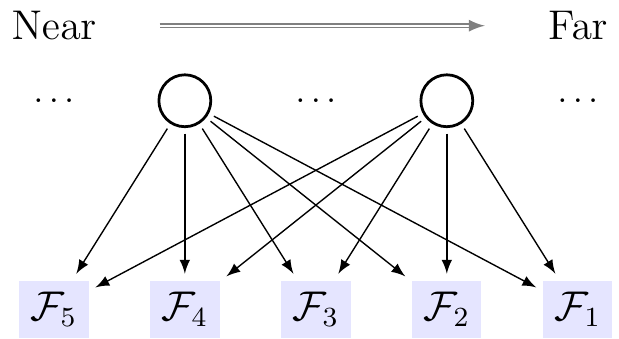}
\begin{small}
\begin{tabular}{C{0.45\linewidth}C{0.45\linewidth}}
(a) Standard strategy~\cite{Lin17e} & (b) Proposed strategy \\
\end{tabular}
\end{small}
\vspace{-6mm}
\caption{{\bf Sampling strategies during training.} 
    Let the circles denote all the training instances sorted in increasing order of depth from left to right. {\bf (a)} The traditional sampling strategy assigns each instance to a single pyramid level according to its size during training. For example, the red instance is fed only to pyramid level ${\cal F}_2$, thus encouraging only this level to yield a reasonable prediction for this sample. {\bf (b)} We propose to assign each instance to multiple pyramid levels, encouraging every pyramid level to produce a reasonable pose estimate for every instance.
    % {\bf (a)} The traditional sampling strategy assigns them to fixed pyramid levels based on thresholds related to the object size~\cite{Lin17e}. {\bf (b)} We propose to make different levels interactivable\WJ{(word does not exist)} during training and make the multi-scale architecture ensemble-ready for robust 6D pose estimation.\YH{Fixed}
    }
\label{fig:sampling_demo}
\end{figure}

\subsection{Ensemble-Aware Sampling}
\label{sec:ea_sampling}

% In the object detection literature, the standard approach to training an FPN is to assign each training sample to a single feature level according to object size. While this process might be sufficient when using a set of pre-defined anchor boxes for detection, it is sub-optimal in our 6D pose estimation context that does not rely on such anchors, because it does not train each level to provide valid predictions for objects of arbitrary sizes, which can be a problem at test time, where the object scale is unknown.  \MS{Does this sound like a credible explanation?} Fig.~\ref{fig:sampling_demo} illustrates this situation. \PF{It really does not.}

% \yh{
Large-scale variations impose drastic difficulties on the network for accurate prediction for every scale. The standard approach to training an FPN follows a divide-and-conquer strategy, consisting of dividing the whole training set of instances into several non-overlapping groups according to the object size and then assigning different groups to different pyramid levels during training, as illustrated in Fig.~\ref{fig:sampling_demo}(a). This simple strategy may be sufficient for object detection where one can simply choose level producing the best prediction based on the objectness scores during testing. However, for 6D pose estimation, it prevents one from leveraging the predictions of the multiple levels jointly to improve robustness, because, for a given scale, most levels will yield highly noisy estimates as they weren't trained for objects at that scale.
%but, prevents possible cooperation between different pyramid levels as the nonoverlapped training strategy, which we will show that is critical for accurate 6D object pose estimation.
% }\YH{Clearer?}

\begin{figure}[t]
    \begin{center}
    \includegraphics[width=0.49\linewidth]{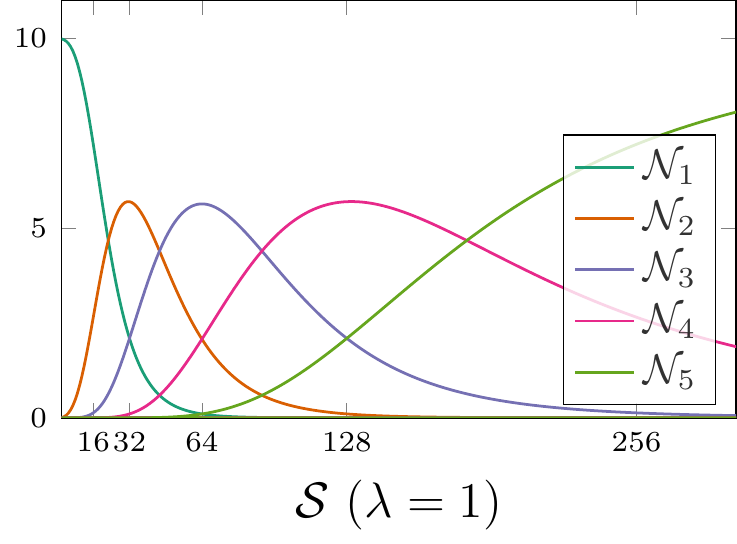}
    \includegraphics[width=0.49\linewidth]{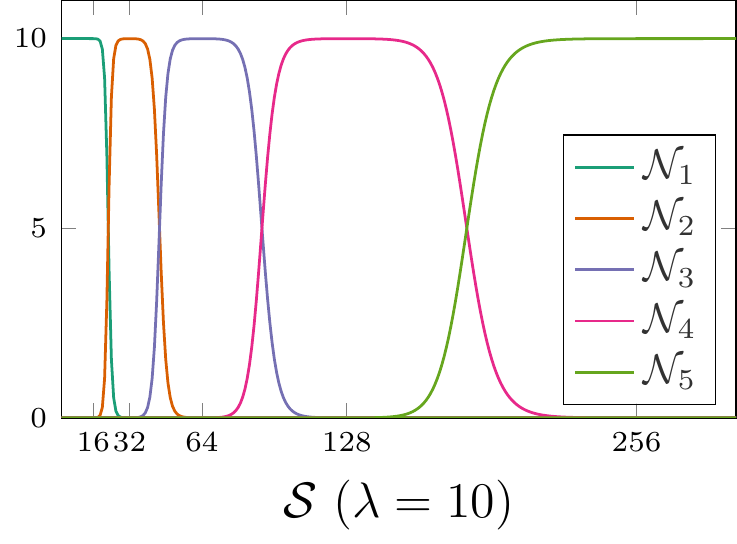}
    % \fbox{\rule{0pt}{2in} \rule{0.25\linewidth}{0pt}}
    \end{center}
    \vspace{-7mm}
    \caption{{\bf Sample count ${\cal N}_k$  at each pyramid level as a function of the object size ${\cal S}$.}
    Typically, when $\lambda > 20$, ${\cal N}_k$ degenerates to the simple ``hard'' assignment strategy that FPN adopts. Note that for a given object size, multiple ${\cal N}_k$s are non-zero, which translates to soft assignments to the different pyramid levels.
    % of ${\cal N}_k$ becomes ``soft'' across multiple pyramid levels in the left figure.
    }
    \label{fig:typical_k_drawings}
\end{figure}

To address this issue, we design a sampling strategy that allows every feature vector within the object segmentation mask at each level to participate in the prediction with a certain probability, as in Fig.~\ref{fig:sampling_demo}(b). 
Let $s_{k}, \text{for } 1\leq k \leq 5$, be a reference object size for level $k$ of the pyramid, chosen based on the object size distribution in the target dataset. For example, in our SwissCube dataset,
% \WJ{first time that CubeSat is mentioned, perhaps just "dataset"?} \YH{Fixed}
we take $s_k$  to be $16, 32, 64, 128$, and $256$, respectively. Then, for an object of size ${\cal S}$ taken to be the largest of the width and height of its 2D bounding box, we uniformly randomly sample
\begin{equation}
    {\cal N}_k=\alpha \frac{e^{-\lambda \Delta_{k}^{2}}}{\sum^{5}_{j=1}{e^{-\lambda \Delta_{j}^{2}}}}\;
\label{eq:nk}
\end{equation}
feature vectors at level $k$ among those within the object segmentation mask, with 
\begin{equation}
    \Delta_k= | \log_{2}\frac{{\cal S}}{s_k} | \; \mbox{ and } \; \alpha=10 \; .
\label{eq:dk}
\end{equation}
The hyper-parameter $\alpha$ specifies the maximum number of active feature vectors on any level, and $\lambda \geq 0$ controls the distribution of the number of active cells across levels. When $\lambda = 0$, all ${\cal N}_k$s are equal, thus using the same number of feature cells at each pyramid level, independently of the object size. By contrast, when $\lambda$ is large, that is, $\lambda > 20$, the sampling strategy degenerates to the ``hard assignment" commonly-used by  FPNs. In Fig.~\ref{fig:typical_k_drawings}, we show how each ${\cal N}_k$ varies as a function of ${\cal S}$ for different $\lambda$ values.  Note that, for a given object size, multiple pyramid levels will be involved in training, thus making them  robust to scale variations.

%and with the corresponding set of feature cells located within segmentation mask on each pyramid level as $C_{k}, 1\leq k \leq 5$, we will choose ${\cal N}_k$ positive feature cells randomly from set $C_k$ for each level $1\leq k \leq 5$. Intuitively, the more similar of the object size ${\cal S}$ as the base interesting size $s_{k}$ of level $k$ the more feature cells should be chosen from, and vice versa. We model it by a normalized function:

%where $\alpha$ and $\lambda$ are adjustable parameters and $\Delta_k$ is the logarithmic scale difference between the current sample and the associated base size of the corresponding pyramid level:

%In practice, we often set $\alpha=10$ which is the total number of activated feature cells on all levels. As to $\lambda \geq 0$ which is the parameter controlling the distribution of the number of activated cells across multiple levels.
%When $\lambda = 0$ all ${\cal N}_k$ will be equal and it will activate the same number of feature cells on each pyramid level. On the other hand, when $\lambda$ is large enough (typically $\lambda \geq 20$) the sampling strategy will degenerate to the simple ``hard'' assignment strategy without any interactions between levels as FPN adopts.
%We show some typical plottings of each ${\cal N}_k$ versus ${\cal S}$ in Fig.~\ref{fig:typical_k_drawings}. Note how the assignment of positive feature cells becomes ``soft'' across multiple pyramid levels, which lays the foundations for our multi-scale fusion presented in the following sections.

% !TEX root = ../top.tex
% !TEX spellcheck = en-US

\subsection{Loss Function in 3D Space}

As mentioned before, every feature vector selected by our sampling procedure is then used to regress the 2D projections of the 8 corners of the 3D object bounding box. %which are obtained offline from the given 3D meshes, where 2D reprojections are represented by the offsets to the center of the corresponding cell~\cite{Hu19a,1,2}. 
When regressing 2D locations, most existing methods~\cite{Rad17,Hu19a} seek to directly minimize the error in the image plane, that is, the loss function $\sum_{i=1}^{n}{|{\bf u}_i-\hat{{\bf u}}_i|}$, where ${\bf u}_i$ is the ground-truth 2D projection and $\hat{{\bf u}}_i$ the predicted one. However, as illustrated by Fig.~\ref{fig:cube_problem}, this loss function is suboptimal, particularly in the presence of large depth variations, because it puts more emphasis on some keypoints than on others and also depends on the object's relative position.
% \WJ{In this section, the frequent use of "keypoint" sounds strange to me. Why not just "point"?}\YH{As points maybe confused with other points on the object surface. We only use the 8 corners of the 3D object bounding box of the object. So we say they are ``key'' points.}

To overcome this, we introduce a loss function in 3D space, which is invariant to the depth of 3D keypoints. Under a perspective camera model, the projection of a 3D object keypoint ${\bf p}_i$ in the image is given by
\begin{equation}
    \begin{aligned}
    \lambda_i
    \begin{bmatrix}
    {\bf u}_i \\
    1 \\
    \end{bmatrix}
    =\bK(\bR{\bf p}_i+{\bf t}),
    \end{aligned}
    \label{eq:perspective}
\end{equation}
where ${\bf u}_i$ is the 2D image location, $\lambda_i$ is a scale factor, $\bK$ is the $3\times 3$ matrix of camera intrinsic parameters, and $\bR$ and ${\bf t}$ are the rotation matrix and translation vector representing the 6D object pose. Then, let
\begin{align}
    \hat{{\bf v}}_i & = {\bf K}^{-1}[\hat{u}_i,\hat{v}_i,1]^\top \\
    {\bf p}_i^c      & = {\bf R}{\bf p}_i+{\bf t}
\end{align}
be the 3D camera ray passing through the predicted 2D location $\hat{{\bf u}}_i = [\hat{u}_i, \hat{v}_i]$ and the corresponding 3D keypoint ${\bf p}_i$ expressed in the camera coordinate system, respectively, where ${\bf R}$ and ${\bf t}$ are the ground-truth rotation matrix and translation vector.  We can then map the re-projection error into 3D space by computing
\begin{equation}
    \begin{split}
    {\bf e}_i   & = {\bf p}_i^c - \hat{{\bf V}}_i {\bf p}_i^c \\
                & = ({\bf I}-\hat{{\bf V}}_i) {\bf p}_i^c\;,
    \end{split}
\end{equation}
where 
\begin{equation}
    \hat{{\bf V}}_i = \frac{\hat{{\bf v}}_i\hat{{\bf v}}_i^\top}{\hat{{\bf v}}_i^\top\hat{{\bf v}}_i}
\end{equation}
is a matrix projecting a 3D point orthogonally to the camera ray $\hat{{\bf v}}_i$~\cite{Lu00}, as illustrated in Fig.~\ref{fig:cube_problem}.
Finally, we take our pose regression loss to be
\begin{equation}
   {\cal L}_{reg} = \sum_{i=1}^{n}{sl_1({\bf e}_i)}.
\end{equation}
where $sl_1(\cdot)$ is the smoothed L1 norm~\cite{Girshick15}. As shown in Fig.~\ref{fig:error_vs_positions},
% \MS{Can you show this in the figure?}\YH{Yes, in a new figure}
this 3D error is consistent across all 3D keypoints and less influenced by the depth and relative position of the observed object. 
% \WJ{"As shown by" sounds odd here, I expected a reference to a figure with data showing an empirical proof that it works, rather than the initial motivation.}\YH{Fixed.}
Furthermore, it can be computed by simple algebraic operations and can thus easily be incorporated in an end-to-end learning formalism.
%Minimizing the reprojection loss measured in 3D space can solve the problem with minimizing 2D reprojection error on the image plane directly as discussed in Fig.~\ref{fig:cube_problem}, making it consistent across different 3D keypoints and also under different 3D locations. There are no complex computations here, only some simple matrix multiplications, which makes it naturally differentiable and can be embedded into the network to let back-propagated gradients update the network weights with ease.

Ultimately, we combine this loss function with that supervising the predicted objectness score, which yields the overall training loss
\begin{equation}
    {\cal L} = \sum_{k=1}^{5}\{{\cal L}_{obj}(k) + {\cal L}_{reg}(k)\},
\end{equation}
where ${\cal L}_{obj}(k)$ and ${\cal L}_{reg}(k)$ are the objectness loss and pose regression loss at level $k$, respectively. In this work, we take the loss ${\cal L}_{obj}$ to be the focal loss~\cite{Lin17f}.

\subsection{Inference via Multi-Scale Fusion}

Thanks to our ensemble-aware sampling strategy, our trained network can produce valid pose estimates at every pyramid level for any test image, independently of its scale. These estimates can be selected by thresholding the objectness score predicted for each feature vector at each level, and in practice we use a threshold $\tau = 0.3$. In principle, these estimates could then be fused directly by a RANSAC+PnP strategy~\cite{Lepetit09} or using the learning-based method of~\cite{Hu20a}. 
For simplicity, we use the RANSAC+PnP approach, but in conjunction with our ensemble-aware sampling scheme.

To apply this scheme at test time, we first need to estimate the object size. To this end, we choose the feature vector leading to the highest objectness score, and compute the size ${\cal S}$ from the corresponding predictions of the 8 bounding box corner projections. Given this size, we then select, for each pyramid level $k$, the ${\cal N}_k$  feature cells that give the highest objectness score. This lets us construct a set of 3D-to-2D correspondences $\{{\bf p}_i \leftrightarrow {\bf u}_{ijk}\}$ for every 3D keypoint ${\bf p}_i$, where ${\bf u}_{ijk}$ is the 2D location predicted for ${\bf p}_i$ by cell ${\cal C}_j$ on feature map ${\cal F}_k$, with $1\le i \le 8, 1\le j \le {\cal N}_k $ and $1\le k \le 5$. Finally, we use a RANSAC based PnP algorithm to obtain a robust 6D pose estimate from these correspondences. We will show in our experiments that this outperforms the prediction obtained from any individual pyramid level.

% \input{fig/swisscube_demo.tex}

% !TEX root = ../top.tex
% !TEX spellcheck = en-US

\section{Experiments}
\label{sec:experiments}

In this section, we first evaluate our framework on the SPEED dataset, and then introduce the SwissCube dataset, which contains accurate 3D mesh and physically-modeled astronomical objects, and perform thorough ablation studies on it. We further show results on real images of the same satellite. Finally, to demonstrate the generality of our approach we evaluate it on the standard Occluded-LINEMOD dataset depicting small depth variations. 
% \WJ{"CubeSat" leaks author nationalities and even institution (it's an EPFL project). In my community, this would be perceived negatively during peer review. I'd suggest using a temporary name ("NanoSat") with an asterisk/footnote saying that the dataset name is temporarily anonymized as not to reveal authorship.} \YH{I am not sure if it is in CV, while it does not hurt to change to a temporary name.} \MS{I agree, but I suggest "CubeSat", which is the standard term for this type of satellite.}

% \yh{
We train our model starting from a backbone pre-trained on ImageNet~\cite{Deng09}, and, for any 6D pose dataset, feed it 3M unique training samples obtained via standard online data augmentation strategies, such as random shift, scale, and rotation. To evaluate the accuracy, we will report the individual performance under different depth ranges, using the standard ADI-0.1d~\cite{Hu19a,Hu20a} accuracy metrics, which encodes the percentage of samples whose 3D reconstruction error is below 10\% of the object diameter. On the SPEED dataset, however, we use a different metric, as we do not have access to the 3D SPEED model, making the computation of ADI impossible. Instead, we use the metric from the competition, that is, ${\bf e}_{\bf q}+{\bf e}_{\bf t}$, where ${\bf e}_{\bf q}$ is the angular error between the ground-truth quaternion and the predicted one, and ${\bf e}_{\bf t}$ is the normalized translation error. Furthermore, because the depth distribution of SPEED is not uniform, with only few images depicting the satellite at a large distance from the camera, we only report the average error on the whole test set, as in the competition.
% }
The source code and dataset are publicly available at \href{https://github.com/cvlab-epfl/wide-depth-range-pose}{https://github.com/cvlab-epfl/wide-depth-range-pose}.

\subsection{Evaluation on the SPEED Dataset}
Although the SPEED dataset has several drawbacks, discussed in Section~\ref{sec:related}, it remains a valuable benchmark, and we thus begin by evaluating our method on it. As the test annotations are not publicly available, and the competition is not ongoing, we divide the training set into two parts, 10K images for training and the remaining 2K ones for testing.
We evaluate the two top-performing methods from the competition,~\cite{Chen19DLR} (DLR) 
% \MS{Don't they have a better name?}
and~\cite{Hu19a} (SegDriven-Z), on these new splits using the publicly-available code, and find their errors to be of similar magnitude to the ones reported online during the challenge.
Note that our method, as DLR and SegDriven-Z, uses the 3D model to define the keypoints whose image location we predict. We therefore exploit a method of~\cite{Hartley00} to first reconstruct the satellite from the dataset. 

Table~\ref{tab:speed_stoa} compares our results to those of the two top-performing methods on this dataset. Note that DLR combines the results of 6 pose estimation networks, followed by an additional pose refinement strategy to improve accuracy. We therefore also report the results of our method with and without this pose refinement strategy. Note, however, that we still use a single pose estimation network. Furthermore, for our method, we report the results of two separate networks trained at different input resolutions. 
At the resolution of 960$\times$, we outperform the two state-of-the-art methods, while our architecture is much smaller and much faster. To further speed up our approach, we train a network at a third (640$\times$) of the raw image resolution. This network remains on par with DLR but runs 20+ times faster.
% \MS{What resolution does Chen use? If they use 960, I would tend to turn this the other way around: Say that, at the same resolution, we outperform the two state-of-the-art methods, while our architecture is much smaller and much faster. To further speed up our approach, we train a network at a third of the raw image resolution. This network remains on par with Chen but runs 50 times faster.}
% \YH{As they use two networks, we can not compare the resolution directly. In more detail. they train a 768x768 detector first and resize all the detected bounding box to 768x768 again to feed into the next pose network. And they use the second pose network 6 times for results ensemble.}
% The faster version can already on par with the top performer but runs 10+ times faster. Our slower version performs the best and still runs 5+ times faster than the competitors.
% \WJ{Can you explain the rationale for these versions with different resolutions? Just speed? Wasn't clear from the text.}\YH{Yes, mainly for speed, also for GPU memory consumption, especially during training.}
% \WJ{You could more prominently point out speed as one of the benefits in the introduction. Practical usage in an autonomous satellite will require low-latency low-compute answers.}\YH{Fixed}

% ~\footnote{\href{https://github.com/BoChenYS/satellite-pose-estimation}{https://github.com/BoChenYS/satellite-pose-estimation}}$^{,}$\footnote{\href{https://github.com/cvlab-epfl/segmentation-driven-pose}{https://github.com/cvlab-epfl/segmentation-driven-pose}}

% \input{fig/swisscube_statistics.tex}

% !TEX root = ../top.tex
% !TEX spellcheck = en-US

\begin{table}
    \centering
    \scalebox{0.8}{
    \begin{small}
    % \rowcolors{2}{white}{gray!10}
    \begin{tabular}{cccccc}
        \toprule
        &&	\multicolumn{2}{c}{Accuracy} & \multirow{2}{*}{Model Size} & \multirow{2}{*}{FPS}\\
        && Raw & Refinement & \\
        \midrule
        \multicolumn{2}{l}{SegDriven-Z~\cite{Hu19a}} & 0.022 & - & 89.2 M & 3.1 \\
        \multicolumn{2}{l}{DLR~\cite{Chen19DLR}} & 0.017 & 0.012 & 176.2 M & 0.7 \\
        \midrule
        \multirow{2}{*}{\bf Ours} 
        & 640$\times$ & 0.018 & 0.013 & {\bf 51.5 M} & {\bf 35} \\
        & 960$\times$ & {\bf 0.016} & {\bf 0.010} & {\bf 51.5 M} & 18 \\
        % {\bf Ours} & {\bf 51.5 M} & {\bf $\sim$ 45 ms} \\
        % Chen {\it etc.} & 48.4 + 21.3 $\times$ 6 = 176.2 M & $\sim$ 1500 ms \\
        % SegDriven-Z & 44.6 + 44.6 = 89.2 M & $\sim$ 300 ms \\ 
        % \midrule
        \bottomrule
    \end{tabular}
    \end{small}
    }
    \vspace{-3mm}
    \caption{{\bf Comparison with the state of the art on SPEED.} Our method outperforms the two top-performing methods in the challenge and is much faster and lighter.}
    \label{tab:speed_stoa}
\end{table} 
\subsection{Evaluation on the SwissCube Dataset}
To facilitate the evaluation of 6D object pose estimation methods in the wide-depth-range scenario, we
introduce a novel SwissCube dataset. The renderings in this dataset account for the
precise 3D shape of the satellite and include realistic models of the star backdrop, Sun, Earth,
and target satellite, including the effects of global illumination, mainly
glossy reflection of the Sun and Earth from the satellite's surface.
To create the 3D model of the SwissCube, we modeled every mechanical part from
raw CAD files, including solar panels, antennas, and screws, and we
carefully assigned material parameters to each part.

The renderings feature a space environment based on the relative placement and
sizes of the Earth and Sun. Correct modeling of the Earth is most important, as
it is often directly observed in the images and significantly affects the
appearance of the satellite via inter-reflection. We extract a high-resolution
spectral texture of the Earth's surface and atmosphere from published data
products acquired by the NASA Visible Infrared Imaging Radiometer Suite (VIIRS)
instrument. These images account for typical cloud coverage and provide
accurate spectral color information on 6 wavelength bands. Illumination from
the Sun is also modeled spectrally using the extraterrestrial solar irradiance
spectrum. The spectral simulation performed using the open source Mitsuba 2
renderer~\cite{Nimier19} finally produces an RGB output that
can be ingested by standard computer vision tools.

% !TEX root = ../top.tex
% !TEX spellcheck = en-US

\begin{figure}[t]
    \begin{center}
    \includegraphics[width=0.6\linewidth]{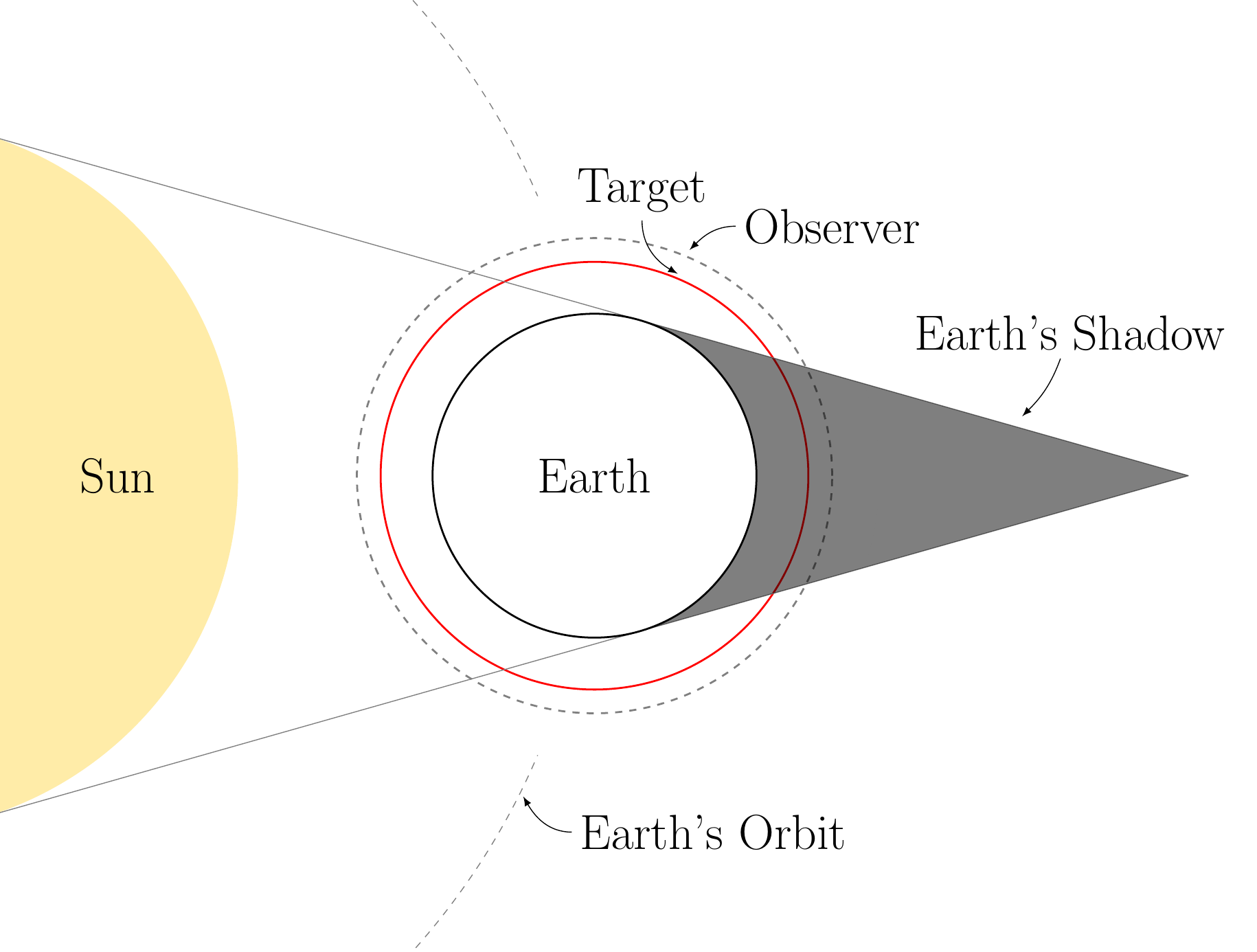}
    % \fbox{\rule{0pt}{2in} \rule{0.25\linewidth}{0pt}}
    \end{center}
    \vspace{-6mm}
    \caption{{\bf Settings for physical rendering of SwissCube.} We physically model the Sun, the Earth, and the complex illumination conditions that can occur in space.}
    \label{fig:render_setting}
\end{figure}

The renderings also include a backdrop of galaxies, nebulae, and star clusters
based on the HYG database star catalog~\cite{hygdatabase} containing around
120K astronomical objects along with information about position and
brightness. The irradiance due to astronomical objects is orders of magnitude
below that of the Sun. To increase the diversity of the dataset, and to ensure
that the network ultimately learns to ignore such details, we boost the
brightness of astronomical objects in renderings to make them more apparent.

Following these steps, we place the SwissCube into its actual orbit located
approximately 700 km above the Earth's surface along with a virtual observer
positioned in a slightly elevated orbit. We render sequences with different
relative velocities, distances and angles. To this end, we use a wide field-of-view (100$^{\circ}$) camera whose distance to the target ranges uniformly between $1d$ to $10d$, where $d$ indicates the diameter of the SwissCube without taking the antennas into accounts.
% \MS{Do you use a wide field-of-view camera? With what angle? Does the diameter $d$ include the antennas, or is it just the cube edge length?}
% \YH{Yes, we use the virtual camera with a FOV of 100. the diameter is computed from only the cube body and does not take the antennas into accounts. And, we treat the Swisscube as an asymmetrical object.}
The high-level
setup is illustrated in Fig.~\ref{fig:render_setting}. Note that the renderings
are essentially black when the SwissCube passes into the earth's shadow, and we
detect and remove such configurations.

\begin{figure*}[t]
    \begin{center}
    % \fbox{\rule{0pt}{1in} \rule{0.25\linewidth}{0pt}}
    \includegraphics[width=0.135\linewidth]{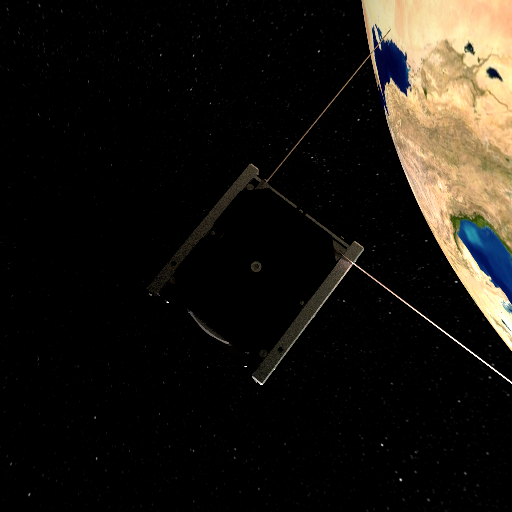}
    \includegraphics[width=0.135\linewidth]{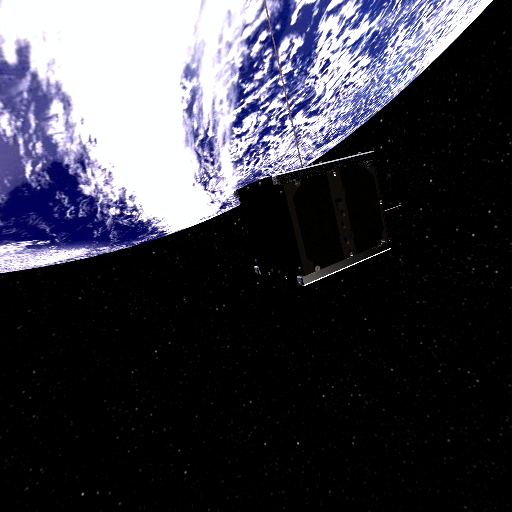}
    \includegraphics[width=0.135\linewidth]{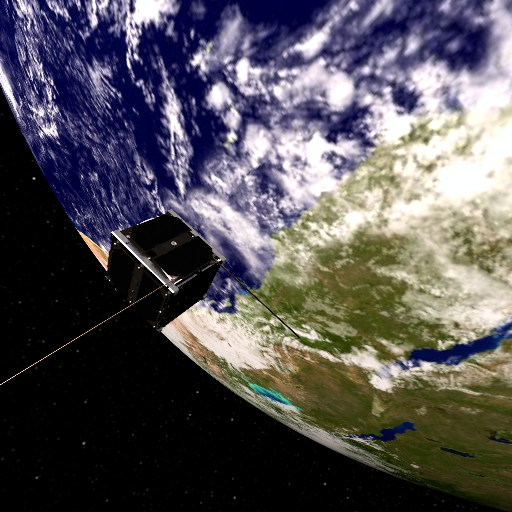}
    \includegraphics[width=0.135\linewidth]{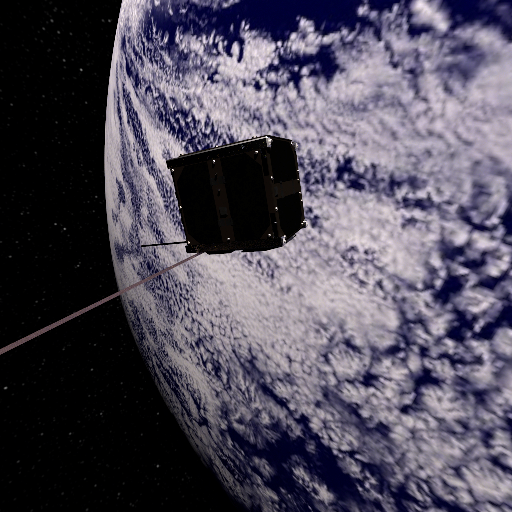}
    \includegraphics[width=0.135\linewidth]{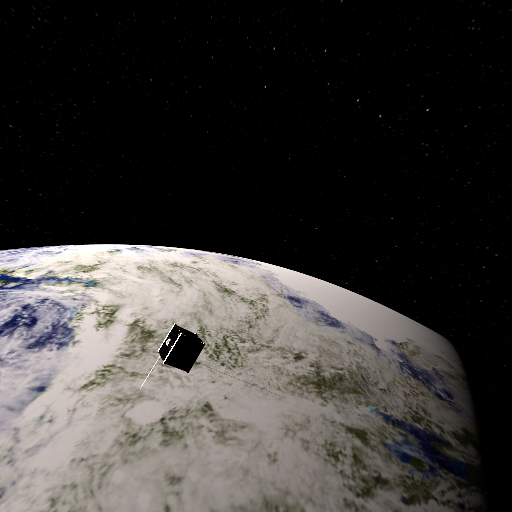}
    \includegraphics[width=0.135\linewidth]{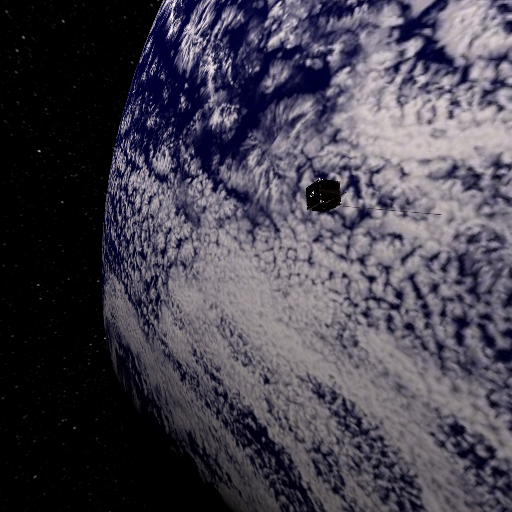} \\
    \includegraphics[width=0.135\linewidth]{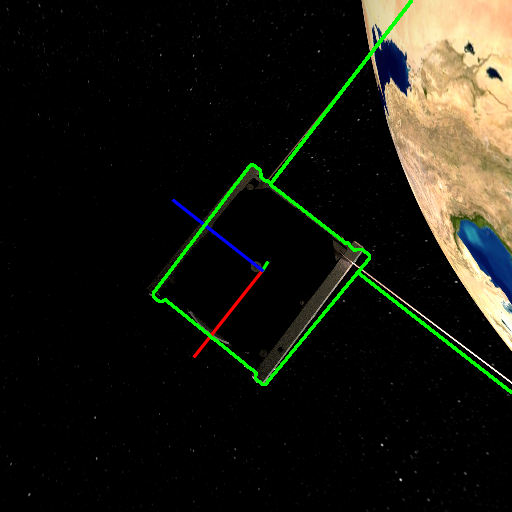}
    \includegraphics[width=0.135\linewidth]{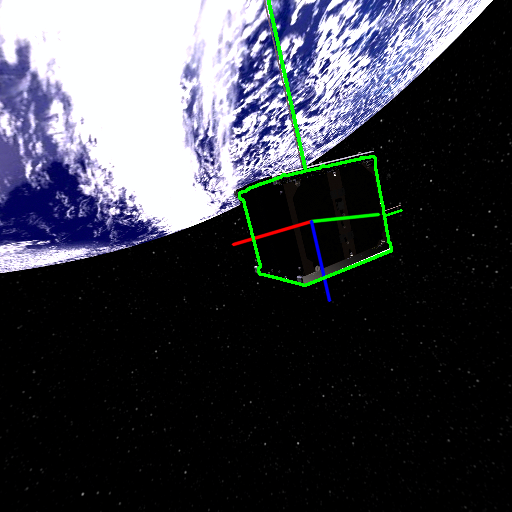}
    \includegraphics[width=0.135\linewidth,trim=0 100 200 100, clip]{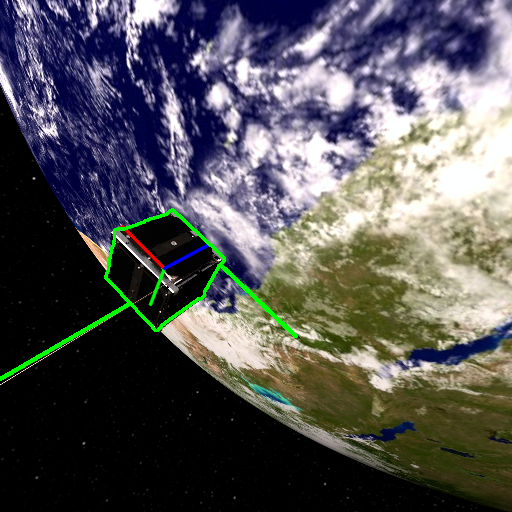}
    \includegraphics[width=0.135\linewidth,trim=0 100 200 100, clip]{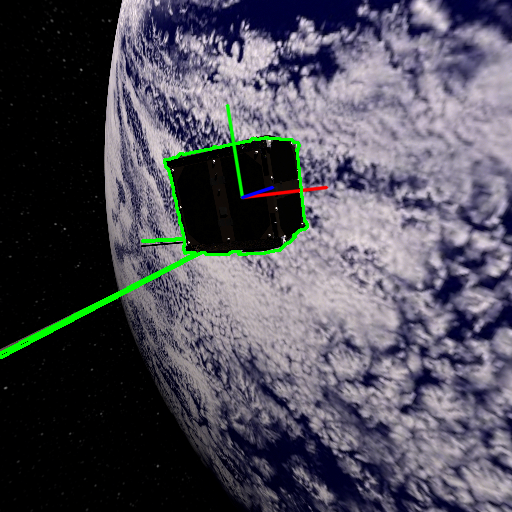}
    \includegraphics[width=0.135\linewidth,trim=100 50 200 250, clip]{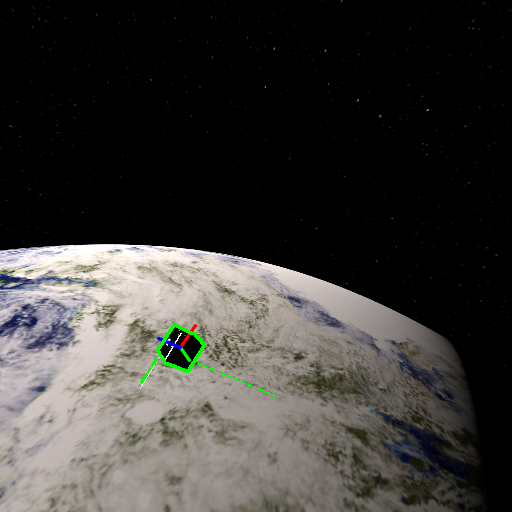}
    \includegraphics[width=0.135\linewidth,trim=250 200 50 100, clip]{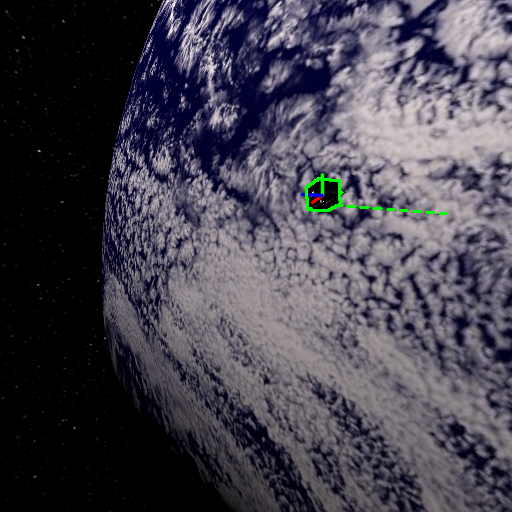}
    \end{center}
    \vspace{-6mm}
    \caption{{\bf Qualitative results on the SwissCube dataset.} Our method yields accurate pose estimates at all scales.}
    \label{fig:results_demo}
\end{figure*}

We generate 500 scenes each consisting of a 100-frame sequence, for a total of
50K images. We take 40K images from 400 scenes for training and the 10K
image from the remaining 100 scenes for testing. 
%We make the depth range of the %CubeSat dataset approximately uniformly distributed from 1d to 10d, as
We render the images at a 1024$\times$1024
resolution, a few of which are shown in Fig.~\ref{fig:results_demo}. During network processing, we resize the
input to 512$\times$512. 
%Although higher input resolution often means higher
%accuracy, as shown by the SPEED experiments, we will focus on this resolution
%setting for a detailed ablation study in this experiment. 
We report the ADI-0.1d accuracy at three
depth ranges, which we refer to as {\it near}, {\it medium}, and {\it far}, corresponding to the depth ranges [1d-4d],
[4d-7d], and [7d-10d], respectively.

% !TEX root = ../top.tex
% !TEX spellcheck = en-US

\subsubsection{Effect of our Ensemble-Aware Sampling}
We first evaluate the effectiveness of our ensemble-aware sampling strategy, further comparing our approach with the single-scale baseline SegDriven~\cite{Hu19a}, which uses the same backbone as us. Note that the original SegDriven method did not rely on a detector to zoom in on the object, but was extended with a YOLOv3~\cite{Redmon18} one in the SPEED competition, resulting in the SegDriven-Z approach evaluated above. For our comparison on the SwissCube dataset to be fair, we therefore also report the results of SegDriven-Z.
% \MS{Could we also evaluate Chen on this dataset? This would be more convincing, although probably too late.} \YH{We had the result, I will add it back.}
Moreover, we also evaluate the top performer on the SPEED dataset, DLR~\cite{Chen19DLR}, on our dataset.

\begin{figure}[t]
    \begin{center}
    \includegraphics[width=0.6\linewidth]{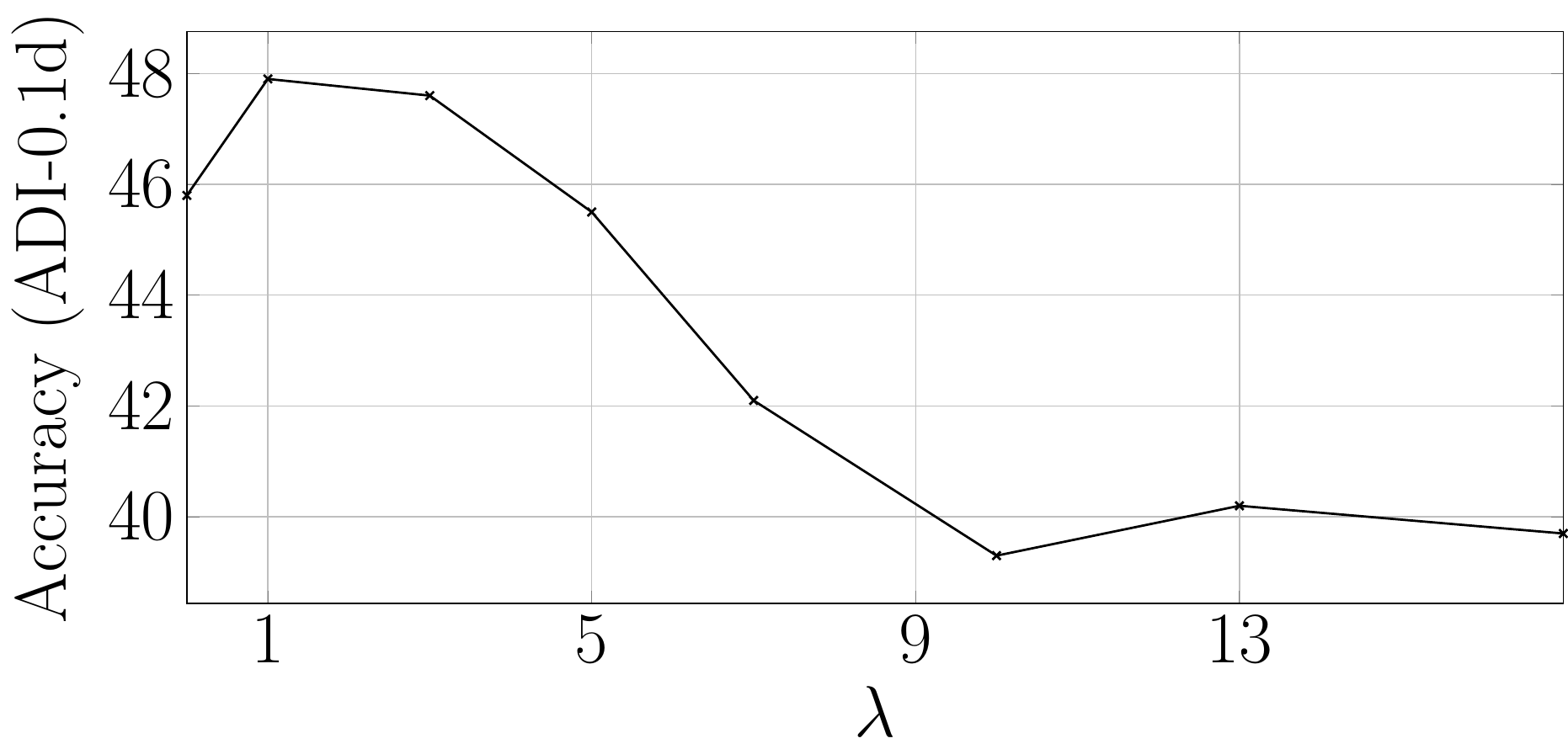}
    % \fbox{\rule{0pt}{2in} \rule{0.25\linewidth}{0pt}}
    \end{center}
    \vspace{-7mm}
    \caption{{\bf Effect of ensemble-aware sampling.} In general, the more cross-level samples are involved in training, that is, the smaller $\lambda$ is, the better the results.}
    \label{fig:param_study}
\end{figure}

Fig.~\ref{fig:param_study} demonstrates the effectiveness of our sampling strategy.
Our results with different $\lambda$ values, which controls the ensemble-aware sampling, show that large values, such as $\lambda>10$, yield lower accuracies. With such large values, our sampling strategy degenerates to the one commonly-used in FPN-based object detectors. This therefore evidences the importance of encouraging every pyramid level to produce valid estimates at more than a single object scale. 
%adopts is much inferior to other settings. That big $\lambda$ makes every pyramid level working on unoverlapped training instances, making different pyramid levels uncombinable during inference for a specific instance. On the other hand, the case of 
Note also that $\lambda=0$, which corresponds to distributing every training instance uniformly to all levels, does not yield the best results, suggesting that forcing every level to produce high-accuracy at all the scales is sub-optimal. In other words, each level should perform well in a reasonable scale range, but these ranges should overlap across the pyramid levels. 
%The imposing of large variation difficulties to every pyramid level makes their performance deteriorate, leading to a worse fusion accuracy. 
This is achieved approximately with $\lambda=1$, which we will use in the following experiments.

Table~\ref{tab:parameters_study} summarizes the comparison results with other baselines. Because it does not explicitly handle scale, SegDriven performs poorly on far objects. This is improved by the detector used in SegDiven-Z. However, the performance of this two-stage approach remains much worse than that of our framework.
Our method outperforms DLR as well, even though our method is 20+ times faster than DLR.
% , independently of the hyper-parameter value $\lambda$, controlling the ensemble-aware sampling. 
Fig.~\ref{fig:results_demo} depicts a few rendered images and corresponding poses estimated with our approach. 
% \MS{I would tend to show this at the end of the first subsection, and potentially compare with SegDriven-Z.}

% !TEX root = ../top.tex
% !TEX spellcheck = en-US

\begin{table}
    \centering
    \scalebox{0.8}{
    \begin{small}
    % \rowcolors{2}{white}{gray!10}
    \begin{tabular}{lcccc}
    \toprule
    & Near & Medium & Far & All  \\
    \midrule
    SegDriven~\cite{Hu19a} &  41.1 & 22.9 & 7.1 & 21.8 \\ 
    SegDriven-Z~\cite{Hu19a} &  52.6 & 45.4 & 29.4 & 43.2 \\ 
    DLR~\cite{Chen19DLR} & 63.8 & 47.8 & 28.9 & 46.8 \\
    {\bf Ours} & {\bf 65.2} & {\bf 48.7} & {\bf 31.9} & {\bf 47.9} \\
    \bottomrule
    \end{tabular}
    \end{small}
    }
    \vspace{-3mm}
    \caption{\bf Our method outperforms all baselines on SwissCube.}
        % Our multi-scale framework outperforms the single-scale baseline SegDriven~\cite{Hu19a} and its zoomed version (SegDriven-Z) significantly, and also DLR~\cite{Chen19DLR}, the top performer on SPEED dataset.
    \label{tab:parameters_study}
\end{table}

% !TEX root = ../top.tex
% !TEX spellcheck = en-US

% !TEX root = ../top.tex
% !TEX spellcheck = en-US

\begin{table}
    \centering
    \scalebox{0.8}{
    \begin{small}
    % \rowcolors{2}{white}{gray!10}
    \begin{tabular}{ccccc}
    \toprule
    & Near & Medium & Far & All \\
    \midrule
    L1 & 0 & 25.2 & \underline{31.8} & 19.5 \\
    L2 & 36.5 & \underline{48.4} & 27.7 & 38.2 \\
    L3 & \underline{62.3} & 47.4 & 19.9 & \underline{42.6} \\
    L4 & 59.2 & 20.2 & 1.7 & 26.3 \\
    L5 & 25.5 & 0.9 & 0 & 8.3 \\
    \midrule
    {\bf Fusion} & {\bf 65.2} & {\bf 48.7} & {\bf 31.9} & {\bf 47.9} \\
    \bottomrule
    \end{tabular}
    \end{small}
    }
    \vspace{-3mm}
    \caption{{\bf Effect of the multi-scale fusion.} Each pyramid level favors a specific depth range, which our multi-scale fusion strategy leverages to outperform every individual level.}
    \label{tab:fusion_effect}
\end{table}

\subsubsection{Effect of our Multi-Scale Fusion}

To better understand the role of each pyramid level during multi-scale fusion, we study the accuracy obtained using the predictions of each individual pyramid level.
Intuitively, we expect the levels with a larger receptive field (feature maps with low spatial resolution) to perform well for close objects, and those with a small receptive field (feature maps with high spatial resolution) to produce better results far-away ones. While the results in Table~\ref{tab:fusion_effect} confirm this intuition for Levels L1, L2 and L3, we observe that the performance degrades at L4 and L5. We believe this to be due to the very low spatial resolution of the corresponding feature maps, 8$\times$8, and 4$\times$4, respectively, making it difficult for these levels to output precise poses. Nevertheless, the accuracy after multi-scale fusion outperforms every individual level, and we leave the study of a different number of pyramid levels to future work.
% \MS{This suggests that we should probably just stop at L3...}\YH{Although the performance of L4 and L5 alone is bad, we are not sure if the L4 or L5 can contribute to the final loss via ensemble. We need more experiments to verify it, so leave it as it is right now.}

%performance with the results combined only from feature cells within each level's segmentation mask. Table~\ref{tab:fusion_effect} shows the results. In intuition, levels with larger reception fields perform better for closer objects and vice versa. However, we find that this is not always true. The performance of level 4 on near objects can not match the one on level 3, and level 5 becomes even more worse. Note that, the spatial feature dimensions for L1, L2, L3, L4, and L5 are 64$\times$64, 32$\times$32, 16$\times$16, 8$\times$8, and 4$\times$4, respectively. Although L4, especially L5, has larger reception fields, the lower spatial resolution makes them less discriminable against 2D keypoints and introduces more visual noises for each cell. Nevertheless, the accuracy after multi-scale fusion outperforms every single level and we leave the study of a different number of pyramid levels to future work.

\subsubsection{Effect of the 3D Loss}

% !TEX root = ../top.tex
% !TEX spellcheck = en-US

\begin{table}
    \centering
    \scalebox{0.8}{
    \begin{small}
    % \rowcolors{2}{white}{gray!10}
    \begin{tabular}{cccccccccc}
    \toprule
    & Near & Medium & Far & All \\
    \midrule
    2D loss & 64.6 & 42.0 & 24.0 & 43.1   \\
    {\bf 3D loss} & {\bf 65.2} & {\bf 48.7} & {\bf 31.9} & {\bf 47.9} \\
    \midrule
    Delta & +0.6 & +6.7 & +7.9 & +4.8 \\
    \bottomrule
    \end{tabular}
    \end{small}
    }
    \vspace{-3mm}
    \caption{{\bf Effect of the 3D loss.} The proposed 3D loss outperforms the 2D one in every depth ranges. The farther the object, the more obvious the advantage of the 3D loss.}
    \label{tab:error_3d_vs_2d}
\end{table} 

% !TEX root = ../top.tex
% !TEX spellcheck = en-US

\begin{figure}[t]
\centering
\includegraphics[width=0.6\linewidth]{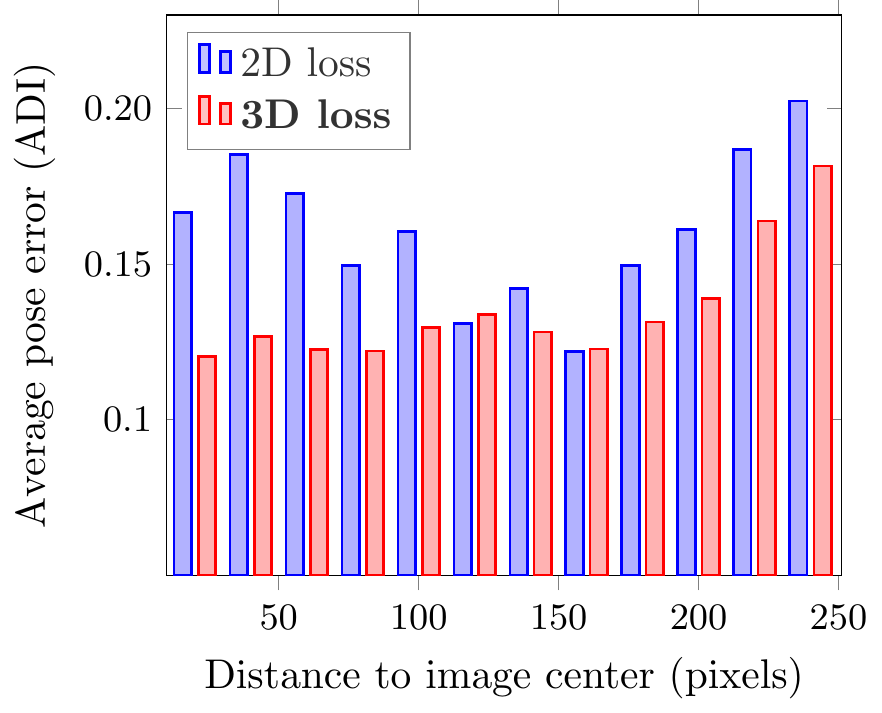}
\vspace{-3mm}
\caption{\small {\bf Pose error as a function of the object position.} The performance of the 2D loss clearly degrades for objects near the image center, whereas that of our 3D loss doesn't. See Fig.~\ref{fig:cube_problem}(b) for the underlying geometry. Note that as the object moves closer to the image boundary, it becomes truncated, which degrades the performance of both losses.}
\label{fig:error_vs_positions}
\end{figure}

%The popular 2D reprojection loss has server problems in the wide-depth-range scenarios as discussed in Fig.~\ref{fig:cube_problem}. To fairly compare the proposed 3D loss against the 2D loss, we train our framework two times from the same initial states and with the same other settings except for the adopted regression loss. 
In Table~\ref{tab:error_3d_vs_2d}, we compare the results obtained by training our approach with either the commonly-used 2D reprojection loss or our loss function in 3D space. Note that our 3D loss outperforms the 2D one in all depth ranges, and the farther the object, the larger the gap between the results of the two loss functions.
In Fig.~\ref{fig:error_vs_positions}, we plot the average accuracy as a function of the object image location. The performance of the 2D loss degrades significantly when the object is located near the image center, whereas the accuracy of our 3D loss remains stable for most object positions. Note that, The reason both of them become worse in the right part of the figure is due to the object truncation by image borders.

\subsection{Results on Real Images}

\begin{figure}[t]
\centering
\includegraphics[width=0.29\linewidth,trim=450 380 400 400,clip]{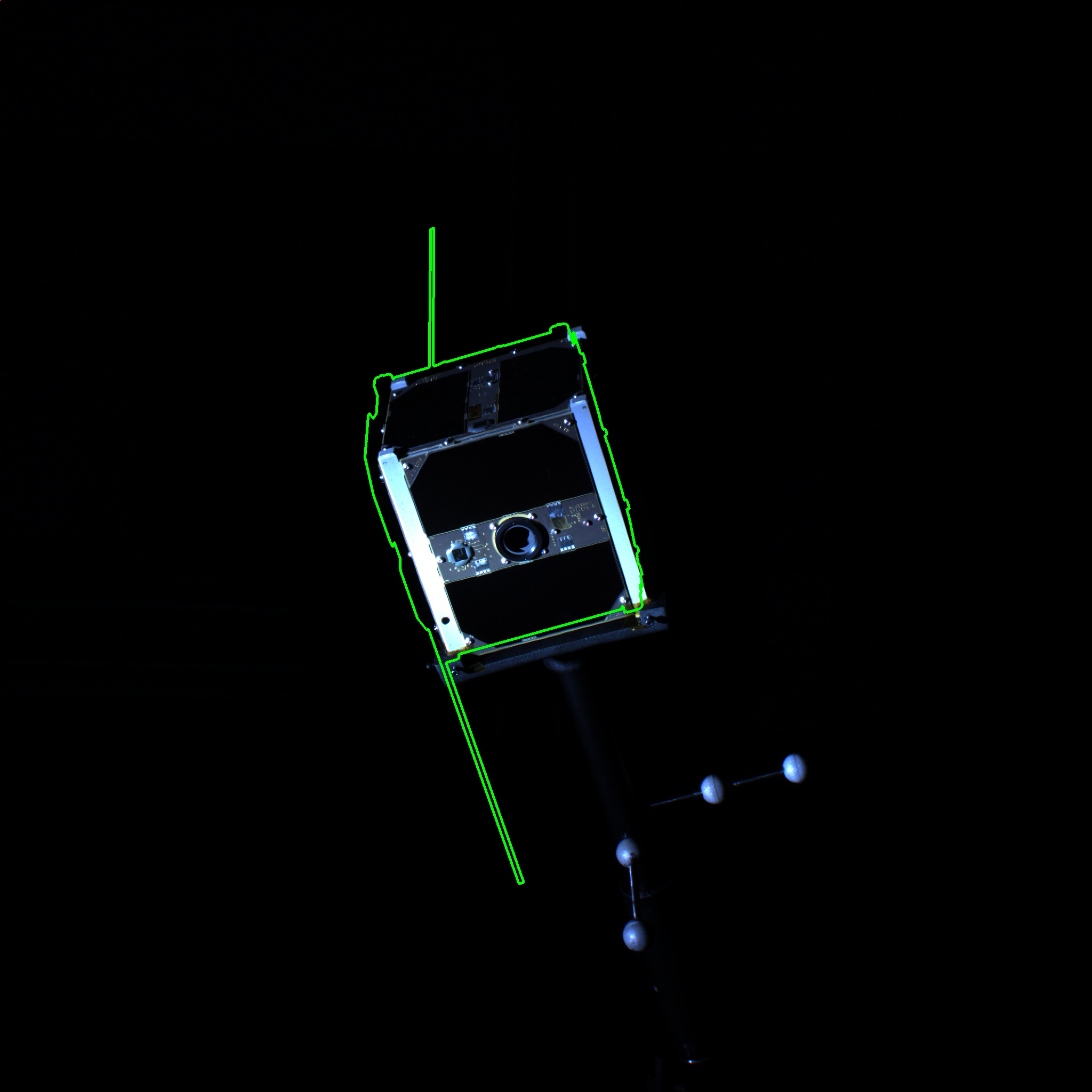}
\includegraphics[width=0.29\linewidth,trim=450 380 400 400,clip]{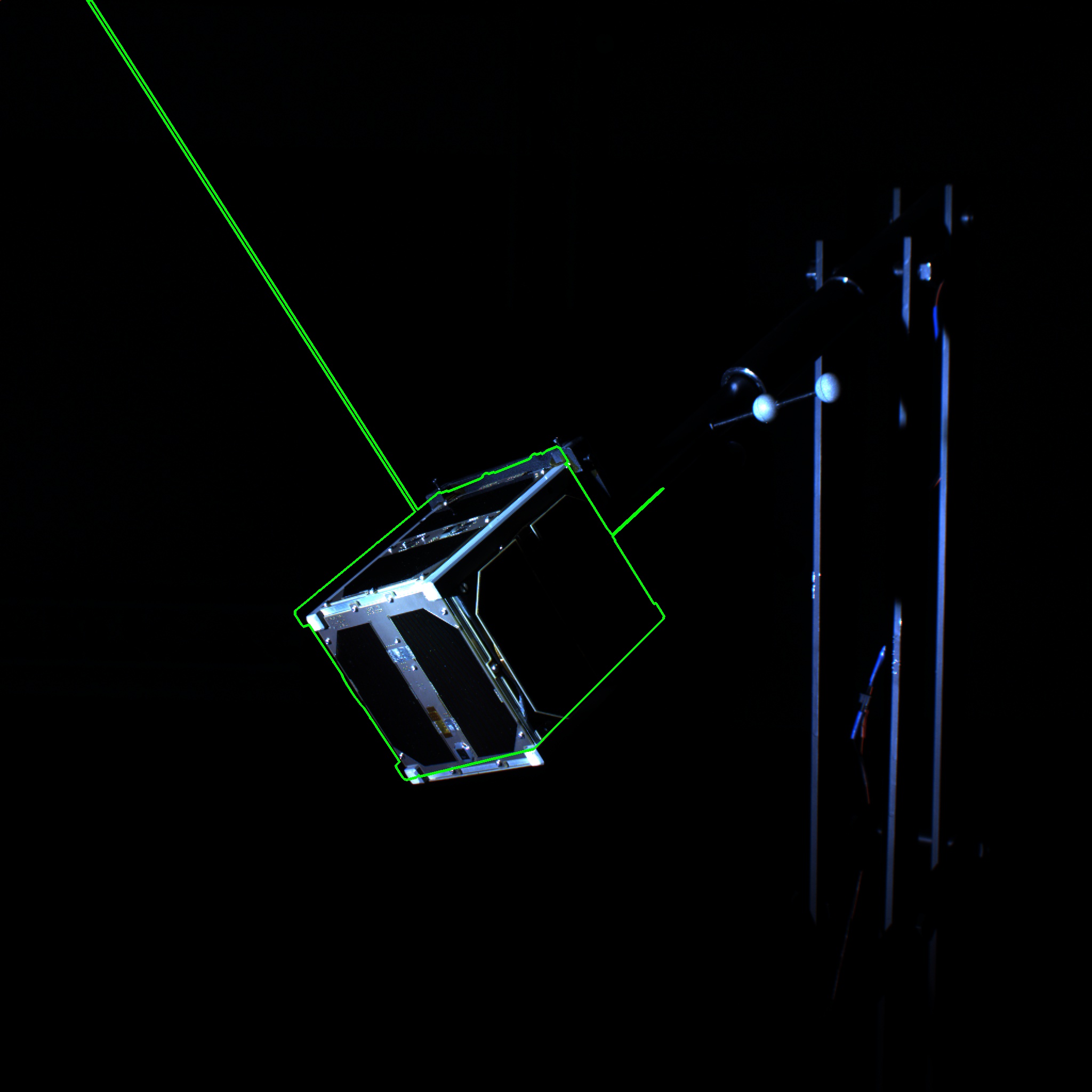}
\includegraphics[width=0.29\linewidth,trim=450 380 400 400,clip]{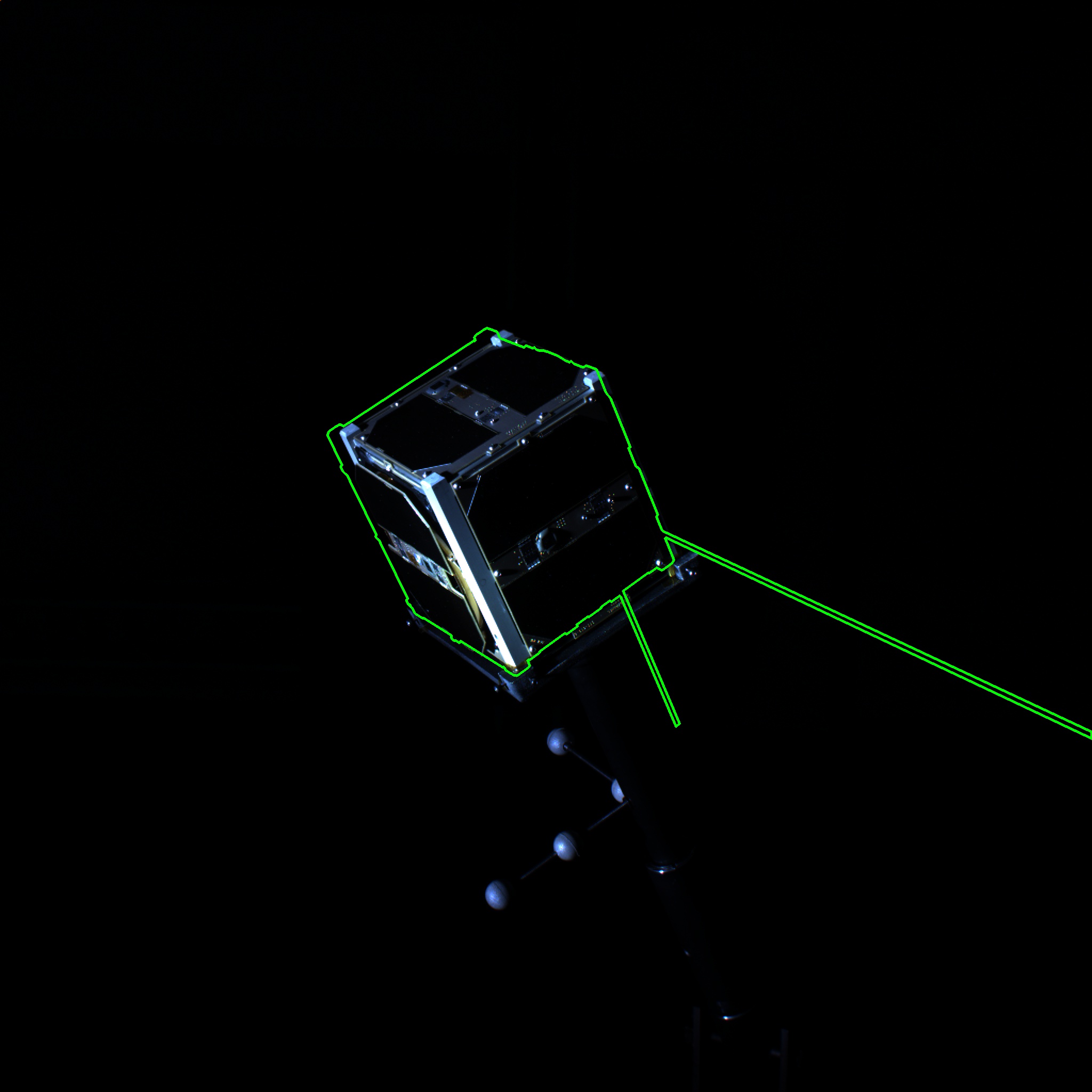}
% \begin{tabular}{cc}
%     \fbox{\rule{0pt}{1.5in} \rule{0.4\linewidth}{0pt}} &
%     \fbox{\rule{0pt}{1.5in} \rule{0.4\linewidth}{0pt}} \\
%     (a) SwissCube & (b) VESPA\\
% \end{tabular}
\vspace{-3mm}
\caption{\small {\bf Qualitative results on real data.} Our model easily adapts to real data, using as few as 20 annotated images.
}
\label{fig:domain_adaptation}
\end{figure}

In Fig.~\ref{fig:domain_adaptation}, we illustrate the performance of our approach on real images. Note that these real images were not captured in space but in a lab environment using a mock-up model of the target and an OptiTrack motion capture system to obtain ground-truth pose information for a few images. We then fine-tuned our model pre-trained on our synthetic SwissCube dataset using only 20 real images with pose annotations. Because this procedure only requires small amounts of annotated real data, it would be applicable in an actual mission, where images can be sent to the ground, annotated manually, and the updated network parameters uploaded back to space.
%Although our CubeSat dataset is rendered by a computer, thanks to its high realism, it can be easily adapted to real data. For the real data, we obtain it by capturing a real-size mock-up of the target. We use a simple finetune~\cite{1}, which is a very basic domain adaptation technique, to adapt our model to the read data. shows some real results on two different satellites, CubeSat and VESPA as well. Although the real data is not captured from the ``real'' space and we are sure we can find better domain adaptation methods, it shines a bright light for the preparation of the real launching in the future.

% \input{table/swisscube_stoa.tex}
% !TEX root = ../top.tex
% !TEX spellcheck = en-US

% !TEX root = ../top.tex
% !TEX spellcheck = en-US

\begin{table}
    \centering
    \scalebox{0.8}{
    % \rowcolors{2}{white}{gray!10}
    \begin{small}
    \begin{tabular}{L{4em}C{3em}C{5em}C{3em}C{3em}}
        \toprule
        & PVNet & SimplePnP & Hybrid & {\bf Ours}\\
        \midrule
        Ape    & 15.8 & 19.2 & 20.9 &  {\bf 22.3} \\
        Can    & 63.3 & 65.1 & 75.3 &  {\bf 77.8} \\
        Cat    & 16.7 & 18.9 & 24.9 &  {\bf 25.1} \\
        Driller& 65.7 & 69.0 & 70.2 &  {\bf 70.6} \\
        Duck   & 25.2 & 25.3 & 27.9 &  {\bf 30.2} \\
    Eggbox$^*$ & 50.2 & 52.0 & 52.4 &  {\bf 52.5} \\
    Glue$^*$   & 49.6 & 51.4 & 53.8 &  {\bf 54.9} \\
       Holepun.& 39.7 & 45.6 & 54.2 &  {\bf 55.6} \\
        \midrule
        Avg.   & 40.8 & 43.3 & 47.5 &  {\bf 48.6} \\
        \bottomrule
    \end{tabular}
    \end{small}
    }
    \vspace{-3mm}
    \caption{{\bf Comparison on Occluded-LINEMOD.} We compare our results with those of PVNet~\cite{Peng19a}, SimplePnP~\cite{Hu20a} and Hybrid~\cite{Song20a}. Symmetry objects are denoted with ``$^*$''.}
    \label{tab:occ_linemod_stoa}
\end{table} 

\subsection{Evaluation on Occluded-LINEMOD}

Finally, to demonstrate that our approach is general, and thus applies to datasets depicting small depth variations, we evaluate it on the standard Occluded-LINEMOD dataset~\cite{Krull15}. Following~\cite{Hu20a}, we use the raw images at resolution 640$\times$480 as input to our network, train our model on the LINEMOD~\cite{Hinterstoisser12b} dataset and test it on Occluded-LINEMOD without overlapped data. Although our framework supports multi-object training, for the evaluation to be fair, we train one model for each object type and compare it with methods not relying on another refinement procedure.
Considering the small depth variations in this dataset, we remove the two pyramid levels with the largest reception fields from our framework, leaving only ${\cal F}_1$, ${\cal F}_2$ and ${\cal F}_3$. As shown in Table~\ref{tab:occ_linemod_stoa}, our model outperforms the state of the art even in this general 6D object pose estimation scenario.

%shows the comparison results of our framework against the state-of-the-art methods. It shows that our multi-scale fusion framework also works pretty well in general 6D object pose estimation.

% !TEX root = ../top.tex
% !TEX spellcheck = en-US

\section{Conclusion}
\label{sec:conclusion}

We have proposed to use a single hierarchical network to estimate the 6D pose of an object subject to large scale variations, as would be the case in a space scenario. Our experiments have evidenced that training the different level of the resulting pyramid for different object scales and fusing their predictions during inference improves accuracy and robustness. We have also introduced the SwissCube dataset, the first satellite dataset with an accurate 3D model, physically-based rendering, and physical simulations of the Sun, the Earth, and the stars. Our approach outperforms the state of the art in both the wide-depth-range scenario and the more classical Occluded-LINEMOD dataset. In the future, we will concentrate on other important aspects of 6D object pose estimation in space, such as removing jitter by 6D pose tracking, and training a usable model with fully-unsupervised real data.

\vspace{0.2em}
{\noindent \bf Acknowledgments.}
This work was supported by the Swiss Innovation Agency (Innosuisse). We would like to thank the EPFL Space Center (eSpace) for the data support.

{\small
\bibliographystyle{ieee_fullname}
\bibliography{string,graphics,vision,learning,space}
}

\end{document}